\documentclass[10pt,twocolumn,letterpaper]{article}

\usepackage{cvpr}              

\usepackage{amsmath,amsfonts,bm,amssymb}
\usepackage{url}
\usepackage{txfonts}
\usepackage[font=small,labelfont=bf,tableposition=top]{caption}
\usepackage{graphicx}
\usepackage{mwe}
\PassOptionsToPackage{table}{xcolor}
\usepackage[table]{xcolor}
\definecolor{blgrey}{rgb}{0.6,0.6,0.6}
\definecolor{bblue}{rgb}{0.855,0.933,0.98}
\definecolor{dblue}{HTML}{5297D6}
\definecolor{gainred}{rgb}{0.1,0.5,0.3}
\definecolor{citecolor}{HTML}{0071BC}
\definecolor{linkcolor}{HTML}{ED1C24}

\newcommand{\gcmark}{\textcolor{green}{\ding{51}}}%
\newcommand{\rxmark}{\textcolor{red}{\ding{55}}}%
\newcommand{\gtriangle}{\textcolor{gray}{$\blacktriangle$}}%

\newrobustcmd*{\mycircle}[1]{\tikz{\filldraw[draw=#1,fill=#1] (0,0) circle [radius=0.1cm];}}

\newrobustcmd*{\mytriangle}[1]{\tikz{\filldraw[draw=#1,fill=#1] (0,0) --
(0.2cm,0) -- (0.1cm,0.2cm);}}

\newrobustcmd*{\myrevtriangle}[1]{%
\tikz{\filldraw[draw=#1,fill=#1] (0,0.2cm) -- (0.2cm,0.2cm) -- (0.1cm,0);}}

\usepackage[normalem]{ulem} 
\usepackage{listings}
\usepackage{color}

\definecolor{dkcyan}{cmyk}{1,0,0,.25}
\definecolor{dkgreen}{rgb}{0,0.6,0}
\definecolor{gray}{rgb}{0.5,0.5,0.5}
\definecolor{mauve}{rgb}{0.58,0,0.82}

\usepackage{tabularx,ragged2e,siunitx}
\newcolumntype{Y}[1]{>{\Centering\hspace{0pt}\hsize=#1\hsize}X}

\usepackage{arydshln}
\definecolor{babyblueeyes}{rgb}{0.63, 0.79, 0.95}
\definecolor{pinegreen}{rgb}{0.0, 0.47, 0.44}
\definecolor{oceanblue}{rgb}{0.0, 0.4, 0.7}
\definecolor{cornellred}{rgb}{0.7, 0.11, 0.11}
\definecolor{yellow}{rgb}{1.0, 0.83, 0.36}
\definecolor{orange}{rgb}{1.0, 0.51, 0.0}
\definecolor{green}{rgb}{0.10, 0.62, 0.47}
\definecolor{violet}{rgb}{0.46, 0.44, 0.70}
\definecolor{pink}{rgb}{0.91, 0.16, 0.54}
\definecolor{codegreen}{rgb}{0,0.6,0}
\definecolor{codeblue}{rgb}{0,0,0.6}
\definecolor{codegray}{rgb}{0.5,0.5,0.5}
\definecolor{codepurple}{rgb}{0.58,0,0.82}
\definecolor{backcolour}{rgb}{0.95,0.95,0.92}
\definecolor{lightblue}{HTML}{84C7F9}
\definecolor{lighterblue}{HTML}{D4ECFF}
\definecolor{lightgreen}{HTML}{b4e5a2}
\definecolor{lightorange}{HTML}{f6c6ad}
\definecolor{myblue}{HTML}{D4ECFF}
\definecolor{mygreen}{HTML}{D0F0C0}
\definecolor{mycham}{HTML}{F7E7CE}
\definecolor{mygray}{gray}{0.90}

\definecolor{teal}{rgb}{0, 0.5, 0.5}

\usepackage[utf8]{inputenc} 
\usepackage[T1]{fontenc}    
\usepackage{booktabs}       
\usepackage{amsfonts}       
\usepackage{nicefrac}       

\newcommand{\black}[1]{\textcolor{black}{#1}}
\newcommand{\cornellred}[1]{\textcolor{cornellred}{#1}}
\usepackage{subcaption}
\usepackage{wrapfig}
\usepackage{tikz}
\usepackage{makecell}
\usepackage{comment}
\usepackage{lmodern}
\DeclareCaptionLabelFormat{andtable}{#1~#2  \&  \tablename~\thetable}

\usepackage{color}
\usepackage{colortbl}
\usepackage{commath}
\usepackage{mathtools}

\usepackage{algpseudocode}
\usepackage{algorithm}
\usepackage{setspace}
\usepackage{mathrsfs}
\usepackage{tcolorbox}
\tcbuselibrary{breakable}
\usepackage{blindtext}
\definecolor{babyblueeyes}{rgb}{0.63, 0.79, 0.95}
\definecolor{pinegreen}{rgb}{0.0, 0.47, 0.44}
\definecolor{oceanblue}{rgb}{0.0, 0.4, 0.7}
\definecolor{cornellred}{rgb}{0.7, 0.11, 0.11}
\definecolor{teal}{rgb}{0, 0.5, 0.5}

\usepackage{multirow}
\usepackage{kotex}
\usepackage{amsmath}

\usepackage[toc,page,header]{appendix}
\usepackage{minitoc}

\usepackage[pagebackref,breaklinks,linkcolor=cornellred,colorlinks=true,citecolor=cvprblue]{hyperref}

\usepackage{pifont}
\usepackage{colortbl}

\def\1{\bm{1}}

\def\vs{{\bm{s}}}
\def\vt{{\bm{t}}}

\def\mA{{\bm{A}}}

\def\mR{{\bm{R}}}

\newcommand{\sigmoid}{\sigma}

\DeclareFixedFont{\ttb}{T1}{txtt}{bx}{n}{12} 
\DeclareFixedFont{\ttm}{T1}{txtt}{m}{n}{12}  

\usepackage{color}
\definecolor{deepblue}{rgb}{0,0,0.5}
\definecolor{deepred}{rgb}{0.6,0,0}
\definecolor{deepgreen}{rgb}{0,0.5,0}

\usepackage{listings}

\newcommand\pythonstyle{\lstset{
language=Python,
basicstyle=\ttm,
morekeywords={self},              
keywordstyle=\ttb\color{deepblue},
emph={MyClass,__init__},          
emphstyle=\ttb\color{deepred},    
stringstyle=\color{deepgreen},
frame=tb,                         
showstringspaces=false
}}

\lstnewenvironment{python}[1][]
{
\pythonstyle
\lstset{#1}
}
{}

\newcommand\pythoninline[1]{{\pythonstyle\lstinline!#1!}}

\definecolor{cvprblue}{rgb}{0.21,0.49,0.74}

\def\paperID{18325} 
\def\confName{CVPR}
\def\confYear{2025}

\usepackage[accsupp]{axessibility}  
\makeatletter
\newcommand{\printfnsymbol}[1]{%
  \textsuperscript{\@fnsymbol{#1}}%
}
\makeatother
\title{Preserve or Modify? Context-Aware Evaluation for Balancing Preservation and Modification in Text-Guided Image Editing}

\author{
Yoonjeon Kim\textsuperscript{1,\thanks{ Equal contribution,\textsuperscript{$\dagger$}~Corresponding Author.}} \quad
Soohyun Ryu\textsuperscript{1,*} \quad
Yeonsung Jung\textsuperscript{1} \quad
Hyunkoo Lee\textsuperscript{1} \quad \\
Joowon Kim\textsuperscript{1} \quad
June Yong Yang\textsuperscript{1} \quad
Jaeryong Hwang\textsuperscript{2} \quad
Eunho Yang\textsuperscript{1,3,$\dagger$}\\
[2mm]
\textsuperscript{1}~KAIST \quad \textsuperscript{2}~Republic of Korea Naval Academy \quad \textsuperscript{3}~AITRICS
\\
[2mm]
\normalsize{
\url{https://augclip.github.io/}
}
}

\begin{document}
\doparttoc
\faketableofcontents

\maketitle

\begin{abstract}
The development of vision-language and generative models has significantly advanced text-guided image editing, which seeks the \textit{preservation} of core elements in the source image while implementing \textit{modifications} based on the target text. However, existing metrics have a \textbf{context-blindness} problem, indiscriminately applying the same evaluation criteria on completely different pairs of source image and target text, biasing towards either modification or preservation. Directional CLIP similarity, the only metric that considers both source image and target text, is also biased towards modification aspects and attends to irrelevant editing regions of the image. We propose \texttt{AugCLIP}, a \textbf{context-aware} metric that adaptively coordinates preservation and modification aspects, depending on the specific context of a given source image and target text. This is done by deriving the CLIP representation of an ideally edited image, that preserves the source image with necessary modifications to align with target text. More specifically, using a multi-modal large language model, \texttt{AugCLIP} augments the textual descriptions of the source and target, then calculates a modification vector through a hyperplane that separates source and target attributes in CLIP space. Extensive experiments on five benchmark datasets, encompassing a diverse range of editing scenarios, show that \texttt{AugCLIP} aligns remarkably well with human evaluation standards, outperforming existing metrics. The code is available at \url{https://github.com/augclip/augclip_eval}.
\end{abstract}

\section{Introduction}

Building on advancements in vision-language models~\citep{clip,li2022blip,geng2023hiclip}, recent generative models \citep{kawar2022imagic, brooks2022instructpix2pix,hertz2022prompt} have been widely utilized as creative tools for image editing via text instructions. Text-guided image editing models enable the modification of images in response to textual guidance, ensuring that changes are aligned with the provided instructions. The primary objective of these models is to apply the necessary \textit{modifications} guided by the target text while \textit{preserving} most of the source image. 

\begin{figure}[t]
    \centering
    \includegraphics[width=\linewidth]{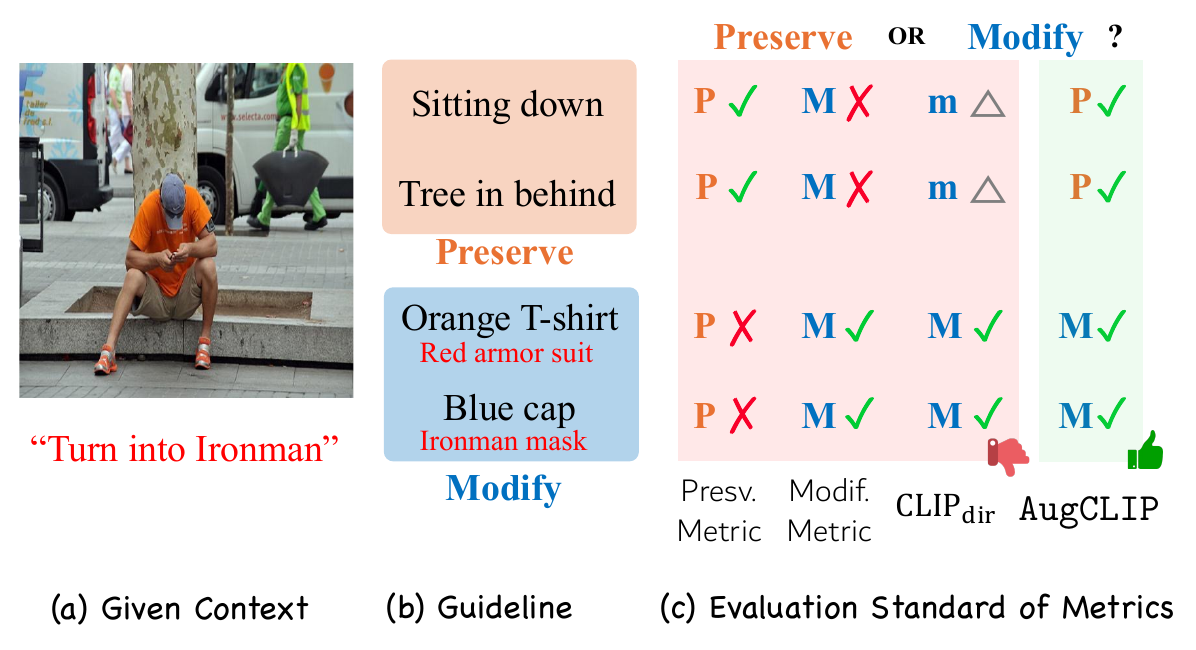}
    \caption{\textbf{Context Blindness Problem of Existing Evaluation Metrics.} Evaluation metrics should consider the specific context of a given source image and target text. However, existing metrics exhibit \textit{context-blindness}, applying the same criteria of either `preserve' (\textbf{\textcolor{orange}{P}}) or `modify' (\textbf{\textcolor{RoyalBlue}{M}}) across the entire image. Our proposed metric, \texttt{AugCLIP}, is a \textit{context-aware} metric that flexibly applies different criteria to local regions of the image.}
    \label{fig:teaser}
\end{figure}

Despite the remarkable advancements in editing models, there has been a lack of rigorous evaluation methods, tailored specifically for text-guided image editing. Consequently, most studies \citep{hertz2023delta, basu2023editval, gal2022stylegan, kim2021diffusionclip, brooks2022instructpix2pix, ruiz2023dreambooth, kocasari2022stylemc} have heavily relied on human evaluation, which provides balanced consideration of preservation and modification aspects. However, as it is costly and impractical for real-world applications, researchers have adapted automatic evaluation metrics~\citep{heusel2017gans, zhang2018unreasonable, caron2021emerging, hessel2021clipscore} originally designed for other vision tasks, such as image generation or captioning. 

In text-guided editing scenarios, it is essential to assess which elements in the source to preserve and which in the target to modify into, based on the specific context of the given source image and target text (see \cref{fig:teaser}\cornellred{(a)-(b)} for example). Existing metrics, however, exhibit \textit{context blindness}, meaning preservation-centric metrics completely ignore the target context while modification-centric metrics do not account for the source context. As shown in \cref{fig:teaser}\cornellred{(c)}, preservation- and modification-centric metrics apply a fixed standard across the entire image, regardless of the true editing requirement of the given context. Even combining them fails to overcome the context blindness problem. 

Among existing metrics, directional CLIP similarity~\citep{gal2022stylegan} is unique because it provides a method to consider both the source image and target text in the evaluation. Even this metric suffers from a context-blindness problem, using a fixed standard without considering the specific relationship between the source image and target text. We show that directional CLIP similarity tends to prioritize modification over preservation, and often focuses on peripheral parts rather than regions that are edited as guided by target text, in \cref{sec:prob_analysis}. These observations underscore the need for a context-aware metric to dynamically adjust the evaluation standard, balancing both aspects in response to diverse editing contexts.

Based on our comprehensive analysis, we propose a novel context-aware metric, \texttt{AugCLIP}, which evaluates the quality of the edited image by comparing it with an estimated representation of a well-edited image. To balance preservation and modification, a well-edited image is represented by a minimum modification on the source image, while matching the target text. We use multi-modal large language models (MLLMs) to extract attributes that capture various visual aspects of the source image and target text. With these attributes, we derive a hyperplane that separates the source and target to derive the ideal modification as a projection to this hyperplane. \texttt{AugCLIP} evaluates how closely the edited image aligns with the estimated ideal image in CLIP space.

Our metric \texttt{AugCLIP} demonstrates remarkable improvement in alignment with human judgment on diverse editing scenarios such as object, attribute, and style alteration compared to all other existing metrics. Moreover, our metric is even applicable to personalized generation, the DreamBooth dataset, where the objective is to identify an object in the source image and edit it into a completely novel context. This shows the flexibility of \texttt{AugCLIP}, which seamlessly applies to a variety of editing directions. Notably, our metric excels in identifying minor differences between the source image and the edited image, showing superb ability in complex image editing scenarios such as MagicBrush.

The major contributions are summarized as follows.
\begin{itemize}
    \item We point out the unreliability of current evaluation metrics in text-guided image editing, noting their inability to balance preservation and modification effectively due to inherent \textbf{context blindness}.
    \item We introduce \texttt{AugCLIP}, a \textbf{context-aware} metric for image editing by automatically \textbf{augmenting} descriptions via MLLM and estimating a balanced representation of preservation and modification, which takes into account the relative importance of each description.
    \item \texttt{AugCLIP} demonstrates a significantly high correlation with human evaluations across various editing scenarios, even in complex applications where existing metrics struggle.
\end{itemize}
\section{Background on Existing Metrics}\label{sec:background}

In the context of text-guided image editing, evaluation metrics should assess the quality of edited images in terms of both preservation of the source image and modification towards the target text. 

\paragraph{Preservation-Centric Metrics} FID \citep{heusel2017gans}, LPIPS \citep{zhang2018unreasonable}, and L2 distance measure feature-level distance between images. These metrics are primarily used in image generation tasks to compare the distributions of real and generated images. When applied to editing scenarios, they assess the distance between the source and edited images, \textit{completely ignoring the target text} guiding the editing process. Segmentation consistency (SC) \citep{kim2021diffusionclip} measures the structural change. Additionally, DINO similarity (DINO) and CLIP-I evaluate the semantic similarity between the source and edited images within the DINO \citep{caron2021emerging} and CLIP \citep{clip} embedding spaces, also focusing solely on preservation.

\paragraph{Modification-Centric Metrics} 
To assess how well an edited image aligns with the target text, the primary approach involves using a pre-trained multi-modal model \citep{clip}. CLIPScore (CLIP-T) \citep{hessel2021clipscore} is the most widely adopted metric in this category, measuring the similarity between the edited image and the target text. This metric \textit{completely ignores the source image} that should be considered for the preservation aspect.

\paragraph{Holistic Evaluation of Preservation and Modification} 
Directional CLIP similarity ($\text{CLIP}_{\text{dir}}$) \citep{gal2022stylegan} is unique among existing metrics as it assesses both preservation and modification aspects by evaluating directional alignment between the original and edited images concerning the source and target text. 

However, these metrics employ a fixed evaluation standard, disregarding the context of each editing scenario, which leads to \textit{context blindness}. As discussed in \cref{sec:prob_analysis}, combining preservation- and modification-centric metrics or using directional CLIP similarity fails to balance preservation and modification adequately, resulting in misalignment with human judgment.

\newcommand{\clipdir}{$\mathrm{CLIP}_\text{dir}$}
\section{Problems of Existing Metrics in Balancing Preservation and Modification Aspects}
\label{sec:prob_analysis}

\begin{figure}[t]
    \centering
    \includegraphics[width=\linewidth]{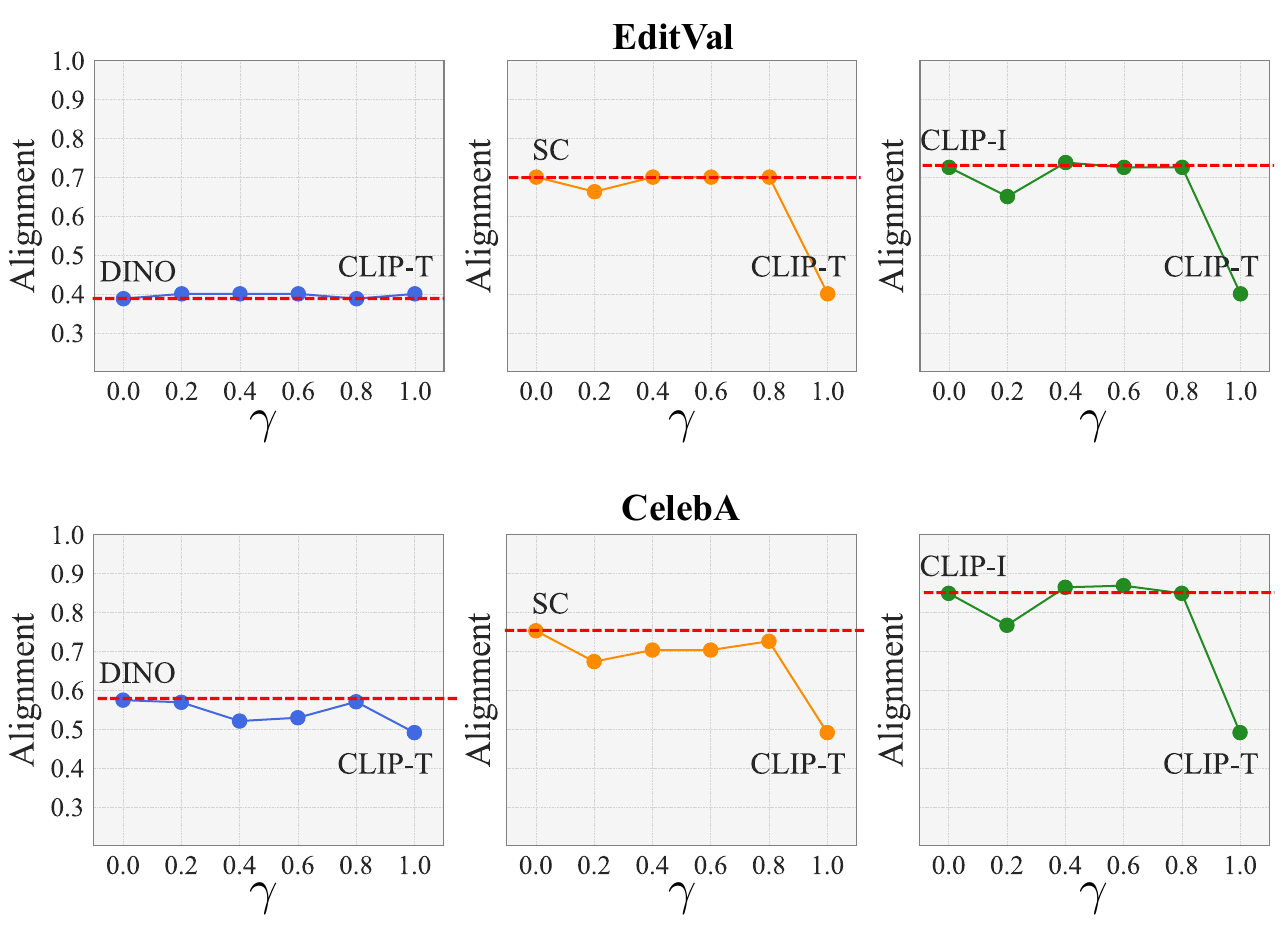}
    \caption{\textbf{Combination of Preservation- and Modification-Centric Metrics Deteriorates in Performance.} The plot shows the human alignment score $\vs_\text{align}$ measured by a linear interpolation of \{DINO, SC, CLIP-I\} and CLIP-T. The results show that combining rather degrades the alignment with human judgment.}
    \label{fig:combine_p_m}
\end{figure}

\begin{figure*}[t]
    \centering
    \includegraphics[width=\linewidth]{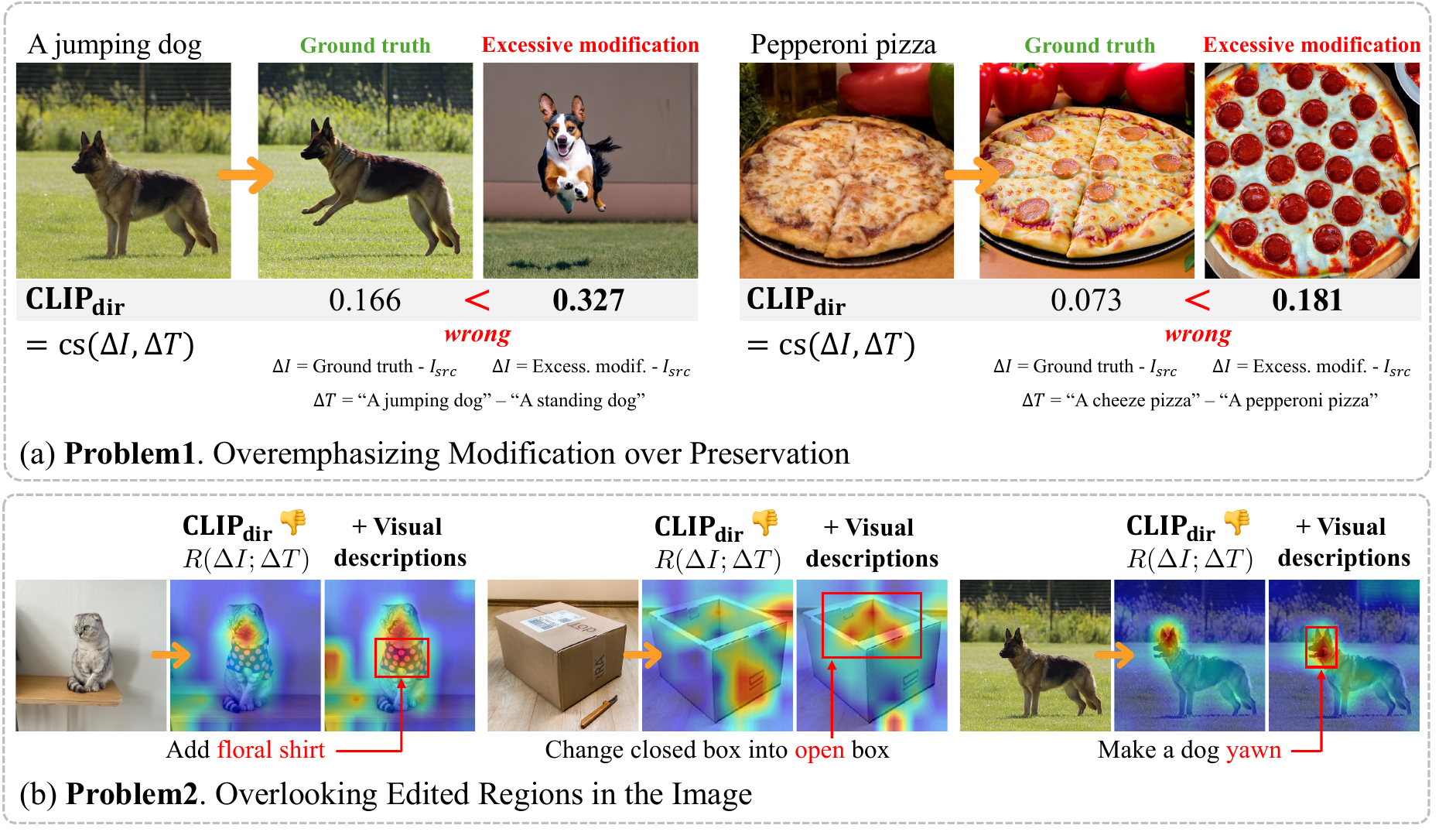}
    \caption{\textbf{Problems of Directional CLIP Similarity.} (a) CLIP$_\text{dir}$ assigns higher scores to excessive modification, over well-edited ground truth images. (b) CLIP$_\text{dir}$ evaluates edited images by attending to irrelevant regions of the image. Adding visual annotations helps \clipdir{}  properly attend to edited regions.}
    \label{fig:clipdir_prob}
\end{figure*}

Existing evaluation metrics, that focus on either one of the preservation or modification aspects, face a challenge in assessing both aspects. Does combining these two metrics successfully balance the preservation and modification aspects? We systematically analyze how these metrics and their combinations still fail in balancing the two colliding objectives, suffering from context-blindness (\cref{subsec:comb_prob}). Moreover, we focus on directional CLIP similarity, the unique metric that accounts for both source image and target text, to reveal that it presents two significant problems that question its reliability (\cref{subsec:clipdir_prob}).

\subsection{Combination of Preservation- and Modification-Centric Metrics}\label{subsec:comb_prob}

We explore whether combining preservation- and modification-centric metrics can improve alignment with human judgments across two benchmark datasets: EditVal \citep{basu2023editval} and CelebA \citep{liu2015faceattributes} (Results on the other three benchmark datasets are deferred to the Appendix). To do so, we combine CLIP-T, a modification-centric metric, and preservation-centric metrics (LPIPS, DINO similarity, CLIP-I) with varying interpolation values $\gamma\in [0,1]$. Note that both metrics are scaled to the same range before interpolation, ensuring the interpolation value $\gamma$ is properly reflected in the final combined score.

As illustrated in \cref{fig:combine_p_m}, the combination of two powerful metrics in preservation and modification often leads to performance degradation. This manifests that a simple combination of the colliding metrics cannot be a reliable evaluation metric. This is due to the context-blindness of these metrics, failing to holistically consider source image and target text. Preservation-centric metrics compare only the source and edited images, ignoring the target context. Similarly, modification-centric metrics consider only the target text, overlooking the source context. This highlights the need for a tailored metric that is contextually aware of both preservation and modification, ensuring a reliable alignment with human evaluations across diverse datasets and tasks.

\subsection{Directional CLIP Similarity}\label{subsec:clipdir_prob}

Directional CLIP similarity (\clipdir{}) is designed to assess both preservation and modification in text-guided image editing by measuring the similarity between image and text directions as follows:
\begin{align} \label{eq:clipdir_def}
  &\text{CLIP}_\text{dir} = \texttt{cs}(\Delta I, \Delta T) \\
  &= \texttt{cs} \Big(E(I_\text{edit}) - E(I_\text{src}), E(T_\text{trg}) - E(T_\text{src}) \Big),
\end{align}
where $\texttt{cs}(\mathbf{a}, \mathbf{b}) = \frac{\mathbf{a} \cdot \mathbf{b}}{\|\mathbf{a}\| \|\mathbf{b}\|}$ denotes cosine similarity and $E(\cdot)$ is a CLIP encoder for either image or text. Here, $I_\text{edit}$ is the edited image to be evaluated, which is edited from the source image $I_\text{src}$ according to the target text $T_\text{trg}$. For evaluation of directional CLIP similarity, $I_\text{src}$, the corresponding text about the source image, is required.
We analyze that \clipdir{} has two major problems that hinder the balanced evaluation of preservation and modification.

\paragraph{\clipdir{} overly emphasizes modification.} Directional CLIP similarity overly emphasizes the alignment with the target text while neglecting the preservation of the source image. Despite attempting to incorporate preservation by subtracting $T_\text{src}$ from $T_\text{trg}$, \clipdir{} frequently favors excessively modified images over ground truth ideally edited images as shown in samples in \cref{fig:clipdir_prob}(a). Specifically, \clipdir{} metric fails on $63\%$ of the 100 editing cases in TEdBench \citep{kawar2022imagic} and $35.1\%$ of the 1053 cases in MagicBrush \citep{zhang2024magicbrush} (Details in \cref{sec:exp}). This problem happens since \clipdir{} is designed to assume that a well-edited image should primarily adhere to the target text. Thus, it blindly favors the visual features of the target text, leading to unreliable results.

\paragraph{\clipdir{} overlooks edited regions.} \clipdir{} often focuses on peripheral or unmodified local areas of the edited image, overlooking the specific areas that are edited by the text. Using the relevancy map \citep{chefer2021generic}, $\mR(\Delta I; \Delta T)$\footnote{The original relevancy map $\mR$ utilizes an image-text pair. To adapt it to \clipdir{}, we subtract two relevancy maps. Details are in the Appendix.}, to visualize attention on image regions corresponding to the text direction, we observe that \clipdir{} frequently fails to concentrate on the image regions that are directly relevant to the target text as shown in \cref{fig:clipdir_prob}(b). The figure also shows that adding explicit visual details guides \clipdir{} to attend to edited regions properly.

\begin{figure*}[t]
    \centering
    \includegraphics[width=\linewidth]{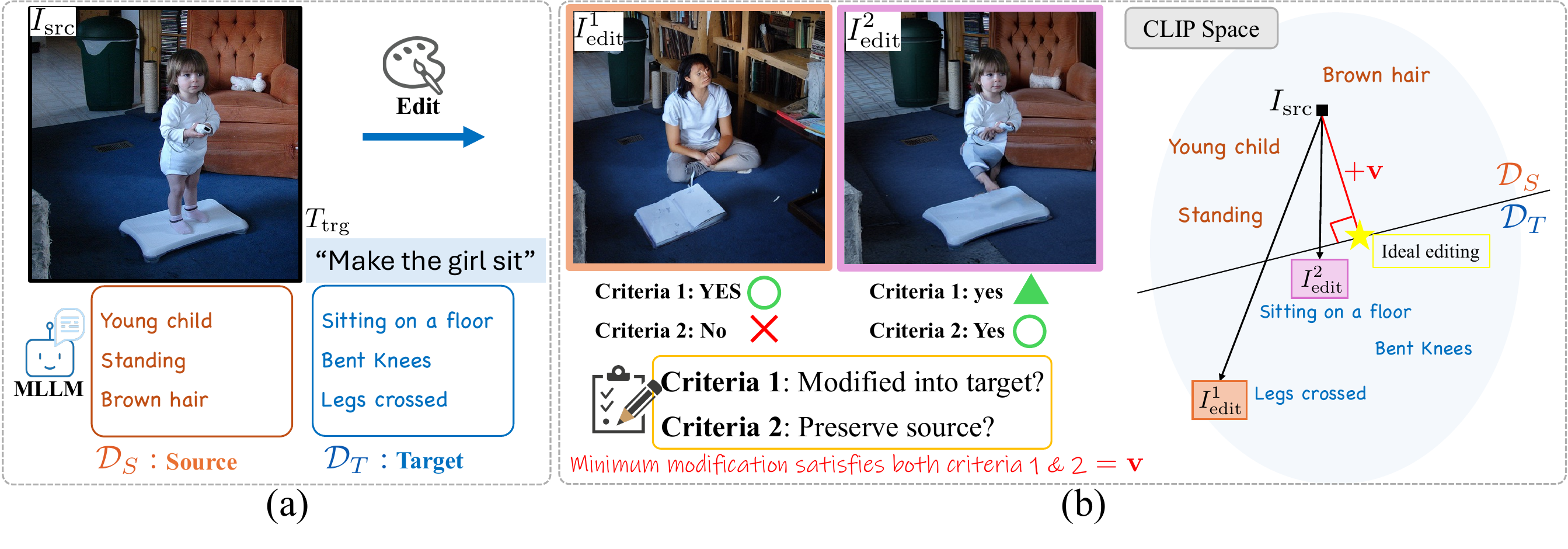}
    \caption{\textbf{(a) Description Extraction Process.} The source image describes a young child standing on the balance board. Target text guides the editing model to make the girl sit. The source and target attributes are extracted with MLLM. \textbf{(b) Evaluation Process of \texttt{AugCLIP}.} The two edited images demonstrate i) An older woman sitting down with legs crossed ($I_\text{edit}^1$) and ii) A young girl sitting on the floor ($I_\text{edit}^2$). \texttt{AugCLIP} derives the ideal image representation as a \textcolor{cornellred}{minimum modification $\mathbf v$} on the source image to be classified as \textit{target}. The second image that is closer to $I_\text{src} + \mathbf v$ shows a higher score, while the first image that is excessively modified with lost source identity shows a lower score.}
    \label{fig:method}
\end{figure*}

\section{AugCLIP: A Context-Aware Evaluation Metric}
\label{sec:augclip}
We introduce a novel evaluation metric, \texttt{AugCLIP}, for scoring the editing quality of the image in terms of preservation and modification. Given a target text $T_\text{trg}$ and a source image $I_\text{src}$, \texttt{AugCLIP} measures the cosine similarity between the edited image $I_\text{edit}$ and the ideal editing. Since the ideal edited image is not given in most evaluation cases, we estimate the representation of an ideally edited image as a modification on the source image, $I_\text{src} + \mathbf v$, where $\mathbf v$ is the modification vector. Then, the editing quality of the edited image $I_\text{edit}$ is evaluated as
\begin{equation}\label{eq:augclip_format}
    \texttt{AugCLIP}\left(I_\text{edit}; I_\text{src}, T_\text{trg}\right) = \texttt{cs}\left(E(I_\text{edit}), E(I_\text{src}) + \mathbf v \right),
\end{equation}
where $E(\cdot)$ is a CLIP encoder.

The evaluation score ranges between $[0, 1]$, where the maximum value $1$ indicates that the edited image is ideally modified by an editing model. We outline the framework of \texttt{AugCLIP} in \cref{fig:method}. We extract the visual features of the source image and target text (\cref{subsec:extract_attr}), then derive the CLIP representation of an ideal modification that satisfies the two criteria: `i) Does the edited image follow the target text?' and `ii) Does the edited image preserve the source image?' (\cref{subsec:key_modif}).

\subsection{Extracting and Processing Attributes}\label{subsec:extract_attr}
\paragraph{Source \& Target Attribute Extraction}
As described in \cref{fig:method}\textcolor{cornellred}{(a)}, we extract visual descriptions about attributes of the source image and target text using a state-of-the-art multi-modal large language model (MLLM), GPT-4V \citep{gpt4v}. From the source image, we extract visual features that can be the object of preservation. Likewise, from the target text, we extract visual features that can be the object of modification. The extraction process is instructed by five annotated in-context examples, which are provided in the Appendix.

\paragraph{Encoding into CLIP Space}
After attribute extraction from MLLM, each attribute is encoded into CLIP space. The source attributes are denoted as $\mathcal D_S$ and target attributes as $\mathcal D_T$.

\paragraph{Relative Weighting on Attributes}
In editing scenarios, some of the attributes are more important than others. For instance, drawing stripes is more essential than a short mane to represent a zebra. To prioritize this, we emphasize the similarity between key attributes within the same class. Since multi-modal large-language models sample based on the empirical probability, frequently extracted attributes indicate higher importance in representing the context. Formally, the importance of each visual attribute in $\mathcal D_S$ and $\mathcal D_T$ is represented as a cosine similarity between CLIP embeddings in the same set as
\begin{equation}
    \textit{Importance}(v_i) = \sum_{j\neq i} \frac{u_i \cdot v_j}{\|u_i\| \|v_j\|},
\end{equation}
for all $u_j$ in the same attribute set, either $\mathcal D_S$ or $\mathcal D_T$.

Additionally, the importance should be down-weighted by how much the visual attribute hinders the preservation of the source image or modification of the target text. More specifically, a collision between source and target attributes is measured by averaged cosine distance to CLIP embeddings in the opposite set as
\begin{equation}
    \textit{Collision}(v_i) = -\sum_{l}\frac{v_i \cdot w_l}{\|v_i\| \|w_l\|},
\end{equation}
for all $w_l$ in the opposite attribute set.

Adding these two measures, the weighting value for the attribute $v \in \mathcal D_S \cup \mathcal D_T$ is defined as 
\begin{equation}\label{eq:weight}
\alpha(v) = \textit{Importance}(v) + \textit{Collision}(v).
\end{equation}
This weighting value accounts for the relative significance of each attribute in the evaluation process and is integrated into \cref{eq:optimization}.

\subsection{Deriving the Ideal Editing in CLIP Space}\label{subsec:key_modif}
Based on the extracted source and target attributes, we describe how to derive a modification vector that satisfies i) modification into the target text and ii) preservation of the source image.

\paragraph{Criterion 1: Does the edited image follow the target text?} To determine if the edited image faithfully follows the given target text, it is intuitive to employ a classifier that classifies the image into either \textit{source} or \textit{target}. Classification into \textit{source} indicates the edited image is still closer to the source image, while classification into \textit{target} means that the image is sufficiently modified. The classifier is derived by a hyperplane that separates the source $\mathcal D_S$ and target $\mathcal D_T$. 

Formally, the classifier function $g(x) = \mathbf w^T x + b$ assigns a CLIP embedding $x$ to the target class if $g(x) > 0$, or to the source class if $g(x) < 0$. Therefore, $E(I_\text{edit})$ should satisfy $\mathbf w^TE(I_\text{edit}) + b > 0$ if the edited image faithfully follows the target text. Here, the optimization form of the classifier function $g(x)$ is
\begin{equation}\label{eq:optimization}
\min_{\mathbf w, b} \frac{1}{2} \|\mathbf w \|^2 + C \sum_{i=1}^N \alpha(v_i) \cdot \max(0, 1 - y_i (\mathbf w \cdot v_i + b)),
\end{equation}
where $y_i = +1$ for $v_i \in \mathcal D_T$ and $y_i=-1$ for $v_i \in \mathcal D_S$. Here, the precomputed weighting values $\alpha(\cdot)$ in \cref{eq:weight} are integrated into the optimization process. 

\paragraph{Criterion 2: Does the edited image preserve the source image?}
However, it is equally important to consider the second criterion, source image preservation. Therefore, the CLIP representation of modification should inflict minimum modification to the original embedding $E(I_\text{src})$, while being classified as a target class. Formally, we derive the modification $\mathbf v$ in CLIP space that satisfies
\begin{equation}\label{eq:condition}
    \underset{\mathbf v}{\text{min}} \|\mathbf v\| \quad \text{subject to} \quad \mathbf w^T \left( E\left(I_\text{src}\right) + \mathbf v\right) + b > 0.
\end{equation}

\paragraph{Derivation of $\mathbf v$}
\black{To minimize $\| \mathbf{v} \|$, $\mathbf{v}$ should be in the direction that most efficiently increases the classifier output $g(\cdot)$. This direction is given by the gradient of the classifier function concerning $\mathbf{x}$, which is $\mathbf{w} = \nabla g(\mathbf{x})$. Therefore, the optimal $\mathbf{v}$ is $c \mathbf{w}$ where $c > 0$ is a scalar to be determined. Substituting $\mathbf{v} = c \mathbf{w}$ to \cref{eq:condition}, we derive the following equation
\begin{equation}
    c > \frac{ - (\mathbf{w}^\top I_\text{src} + b) }{ \| \mathbf{w} \|^2 }.
\end{equation}}
In order to solve for $c$ that minimizes $\| \mathbf{v} \|$, we derive the smallest $c$ satisfying the condition as $c_{\text{min}} = \frac{ - (\mathbf{w}^\top I_\text{src} + b) }{ \| \mathbf{w} \|^2 }$.
Then the modification vector $\mathbf v$ is derived as
\begin{equation}\label{eq:v}
\mathbf v = c_{\text{min}} \mathbf{w} = \frac{ - (\mathbf{w}^\top I_\text{src} + b) }{ \| \mathbf{w} \|^2 } \mathbf{w}.
\end{equation}
Finally, \texttt{AugCLIP} is computed by substituting $\mathbf v$ to \cref{eq:augclip_format}.

\begin{table*}[t]
\centering
\caption{\textbf{Benchmark Datasets and Editing Models Used for Evaluating the Performance of Metrics.}}
\label{tab:dataset_type}
\resizebox{\textwidth}{!}{
        \begin{tabular}{
        l c c c c c
        }
        \toprule
        & CelebA & EditVal & TEdBench & MagicBrush & DreamBooth\\
        \midrule
        \textbf{Dataset Types} & Facial Attribute & General Object & Object Centric & Local Region Editing & Personalized Generation \\
        \textbf{Editing Models} & \citep{kwon2022diffusion,kim2022learning,kim2021diffusionclip,Patashnik_2021_ICCV} &  \citep{hertz2023delta,couairon2022diffedit,brooks2022instructpix2pix,hertz2022prompt,kawar2022imagic} & \citep{hertz2023delta,couairon2022diffedit,brooks2022instructpix2pix,hertz2022prompt,kawar2022imagic} & \citep{hertz2023delta,couairon2022diffedit,brooks2022instructpix2pix,hertz2022prompt,kawar2022imagic} &\citep{li2024blip,kumari2023multi,wei2023elite} \\
        \textbf{Dataset Size} & 50 & 648 & 100 & 1053 & 50 \\
        \bottomrule
        \end{tabular}
        }
\end{table*}

\newcommand{\score}{$\vs_\text{2AFC}$}
\newcommand{\acc}{\textbf{Acc}$_\text{Both}$}
\newcolumntype{P}[1]{>{\centering\arraybackslash}p{#1}}
\newcolumntype{M}[1]{>{\centering\arraybackslash}m{#1}}
\newcolumntype{L}[1]{>{\raggedright\arraybackslash}m{#1}}

\section{Experiments} \label{sec:exp}

\paragraph{Implementation Details.}
For our experiments, we employ a pre-trained CLIP-ViT 16/B model for CLIP-based metrics. Source and target attributes are generated using GPT-4V \citep{gpt4v}. Further details on prompting the source and target attributes are in the Appendix.

\paragraph{Baseline Metrics.} We compare \texttt{AugCLIP} with DINO similarity, LPIPS, L2, Segmentation Consistency (SC), Directional CLIP similarity ($\text{CLIP}_{\text{dir}}$), CLIP-I, and CLIP-T. The CLIP variants are summarized as follows: 
\begin{align*}
    &\text{CLIP}_{\text{dir}} = \texttt{cs}\left(E(I_\text{edit}) - E(I_\text{src}), E(T_\text{trg}) - E(T_\text{src})\right)\\
    &\text{CLIP-I} = \texttt{cs}\left(E(I_\text{edit}), E(I_\text{src})\right)\\
    &\text{CLIP-T} = \texttt{cs}\left(E(I_\text{edit}), E(T_\text{trg})\right).
\end{align*}

Note that FID measures the distance between two distributions of image features, this metric cannot be measured in a sample-wise manner. Therefore, FID cannot be tested in terms of alignment with human evaluation or preference test on ground truth samples.

\paragraph{Benchmark Datasets and Editing Models.}
We evaluate \texttt{AugCLIP} and existing metrics across several text-guided image editing benchmarks, including TEdBench \citep{kawar2022imagic}, EditVal \citep{basu2023editval}, MagicBrush \citep{zhang2024magicbrush}, DreamBooth \citep{ruiz2023dreambooth}, and CelebA \cite{liu2015faceattributes}. \black{Each benchmark dataset represents varying editing scenarios as in \cref{tab:dataset_type}.} For each benchmark dataset, we employ multiple editing models \citep{kwon2022diffusion,kim2022learning,kim2021diffusionclip,Patashnik_2021_ICCV,hertz2023delta,couairon2022diffedit,brooks2022instructpix2pix,hertz2022prompt,li2024blip,kumari2023multi,wei2023elite} as specified in \cref{tab:dataset_type} for suitable type of benchmark dataset. Details on benchmark datasets are provided in the Appendix.

\subsection{Correlation with Human Judgment}
\paragraph{User Study}
To evaluate the effectiveness of \texttt{AugCLIP} compared to existing metrics, we conduct a Two-Alternative Forced Choice survey on human evaluators to collect preferences over diverse edited images. The alignment score \score{} measures if each metric prefers the same option as human evaluators. The survey was conducted using Amazon Mechanical Turk (AMT), over three benchmark datasets, encompassing a total of 105 participants as detailed in \cref{tab:survey_info}. The survey questionnaire is structured as described in the Appendix, providing clear criteria to consider source image preservation and target text alignment. 

\paragraph{\texttt{AugCLIP} aligns better with human judgments.} As shown in \cref{tab:exp_main_2afc}, \texttt{AugCLIP}, the context-aware metric that balances preservation and modification, is superior over all previous metrics across diverse benchmark datasets. A high level of alignment between preservation-centric (L2, LPIPS, DINO, SC, CLIP-I) and human judgment is expected because these benchmarks do not significantly deviate from the source image. Moreover, per-model evaluations are presented alongside human evaluations in \cref{tab:model}. Unlike other metrics, \texttt{AugCLIP} perfectly aligns with human rankings on each dataset.

\begin{table}[t]
    \centering
    \caption{\textbf{User Study Statistics.}}
    \label{tab:survey_info}
    \resizebox{\linewidth}{!}{
        \begin{tabular}{lccc}
            \toprule
            & \textbf{CelebA} & \textbf{EditVal} & \textbf{DreamBooth}\\
            \midrule
            \textbf{Survey questions} & 39 & 35 & 37 \\
            \textbf{Participants} & 45 & 30 & 30 \\
            \bottomrule
        \end{tabular}
    }
\end{table}

\newcommand{\numbercircle}[2]{%
\tikz[baseline=(char.base)]{
    \node[shape=circle, draw=#2, fill=#2!30, text=white, inner sep=1.5pt] (char) {\textbf{#1}};
}}
 
\begin{table}[t]
    \caption{
        \textbf{Correlation with Human Judgment.} Human alignment score \score{} ($\uparrow$), scaled between 0 to 1.
       }
    \label{tab:exp_main_2afc}
    \centering
    \resizebox{\linewidth}{!}{
        \begin{tabular}{
            @{\extracolsep{3pt}}
            L{1.4cm}
            M{0.6cm}
            M{0.6cm}
            @{\hskip 0.2in}
            M{2cm}
            M{2cm}
            M{2cm}
            }
        \toprule
        \textbf{Metrics} & Presv. & Modif. & \textbf{CelebA} & \textbf{EditVal} & \textbf{DreamBooth} \\ \midrule 
        \textbf{L2} & \gcmark & \rxmark & 0.653 & 0.348 & {0.464} \\
        \textbf{LPIPS} & \gcmark & \rxmark & 0.465 & 0.360 & 0.286 \\
        \textbf{DINO} & \gcmark & \rxmark & 0.574 & 0.348 & 0.286 \\
        \textbf{SC} & \gcmark & \rxmark & 0.752 & \underline{0.764} & \underline{0.571} \\
        \textbf{CLIP-I} & \gcmark & \rxmark & \underline{0.848} & 0.730  & \textbf{0.857} \\
        \midrule[0.3pt]
        \textbf{CLIP-T} & \rxmark & \gcmark & 0.491 & 0.399 & 0.321 \\
        \textbf{CLIP$_\text{dir}$} & \gtriangle & \gcmark & {0.673} & 0.697 & 0.357 \\
        \midrule[0.3pt]
        \midrule[0.3pt]
        \large{\texttt{AugCLIP}} & \gcmark & \gcmark & \textbf{0.883} & \textbf{0.831} & \textbf{0.857} \\
        \bottomrule
        \end{tabular}
    }
\end{table}

\begin{table}[h]
\centering
\caption{Rank of various text-guided editing models. Highlighted in \colorbox{green!15}{green} if a metric aligns with human evaluations; otherwise, highlighted in \colorbox{red!15}{red}.}
\vspace{-7pt}
\resizebox{\linewidth}{!}{
\begin{tabular}{@{} clc|ccc|c @{}}
\toprule
\textbf{Dataset}& \textbf{Models} & \textbf{Rank} & \clipdir{} $\uparrow$ & LPIPS $\downarrow$ & \texttt{AugCLIP} $\uparrow$ & \textbf{Human} $\uparrow$ \\ \midrule[1pt]
\multirow{3}{*}{\textbf{DreamBooth}} 
& ELITE & \numbercircle{1}{red} & \cellcolor{red!20} 0.1132 \footnotesize  \numbercircle{2}{green} & \cellcolor{red!20} 71.38 \footnotesize  \numbercircle{2}{green} & \cellcolor{green!20} 0.7642 \footnotesize \numbercircle{1}{red} & 0.8478 \\
& BlipDiffusion & \numbercircle{2}{green} &  \cellcolor{red!20} 0.0836 \footnotesize \numbercircle{3}{blue} & \cellcolor{red!20} 70.88 \footnotesize \numbercircle{1}{red} & \cellcolor{green!20} 0.7579 \footnotesize  \numbercircle{2}{green} &  0.6525 \\
& CustomDiffusion & \numbercircle{3}{blue} & \cellcolor{red!20} 0.1348 \footnotesize \numbercircle{1}{red} & \cellcolor{green!20} 
 73.84 \footnotesize \numbercircle{3}{blue} &  \cellcolor{green!20} 0.6156 \footnotesize \numbercircle{3}{blue} &  0.0263 \\

\midrule
\multirow{3}{*}{\textbf{EditVal}} 
& P2P & \numbercircle{1}{red} & \cellcolor{red!20} 0.1771 \footnotesize \numbercircle{3}{blue} & \cellcolor{green!20} 15.04 \footnotesize \numbercircle{1}{red} & \cellcolor{green!20} 0.8521 \footnotesize \numbercircle{1}{red} & 0.6133 \\
& InstructPix2Pix & \numbercircle{2}{green} &  \cellcolor{green!20} 0.1774 \footnotesize  \numbercircle{2}{green}& \cellcolor{red!20} 25.75 \footnotesize \numbercircle{3}{blue} & \cellcolor{green!20} 0.8242 \footnotesize  \numbercircle{2}{green}  &  0.4855 \\
& DiffEdit & \numbercircle{3}{blue} &\cellcolor{red!20}  0.2272 \footnotesize \numbercircle{1}{red} & \cellcolor{red!20} 20.41 \footnotesize  \numbercircle{2}{green} &  \cellcolor{green!20} 0.8155 \footnotesize \numbercircle{3}{blue} &  0.3214 \\

\midrule
\multirow{3}{*}{\textbf{CelebA}} 
& StyleCLIP &  \numbercircle{1}{red} &  \cellcolor{red!20} 0.0376 \footnotesize  \numbercircle{2}{green} & \cellcolor{green!20} 27.04  \footnotesize \numbercircle{1}{red} & \cellcolor{green!20} 0.8484 \footnotesize \numbercircle{1}{red} & 0.6831 \\
& Multi2One & \numbercircle{2}{green} & \cellcolor{red!20} 0.0414  \footnotesize \numbercircle{1}{red} & \cellcolor{green!20} 27.95 \footnotesize \numbercircle{2}{green} & \cellcolor{green!20} 0.8152 \footnotesize  \numbercircle{2}{green} &  0.5469 \\
& Asyrp & \numbercircle{3}{blue} & \cellcolor{green!20} -0.0001 \footnotesize \numbercircle{3}{blue} & \cellcolor{green!20} 36.98 \footnotesize \numbercircle{3}{blue} & \cellcolor{green!20} 0.7750 \footnotesize \numbercircle{3}{blue} &  0.3197 \\
\bottomrule
\end{tabular}
}
\vspace{-5pt}
\label{tab:model}
\end{table}

\subsection{Ground Truth Editing Selection Test} 
\paragraph{Creating Triplet Dataset} We generate a triplet of images, each representing a well-edited, excessively preserved, and excessively modified image. Among the triplet of images, the well-edited image is provided in the benchmark datasets, TEdBench, and MagicBrush. Excessively modified images are generated with text-to-image models \citep{rombach2022high} instructed by the target text, completely ignoring the source image. Excessively preserved images are generated by applying a negligible transformation to the given source image, ignoring the target text. Then, we measure the accuracy where each metric correctly assigns the highest score to the well-edited image as \acc{}. Yielding high accuracy in this test means that the evaluation metric can balance both preservation and modification aspects, without being biased to either side.

\paragraph{\texttt{AugCLIP} is superior in balancing preservation and modification.} In \cref{tab:exp_main_gt}, \texttt{AugCLIP} is on par with or even better than baseline metrics over all datasets. This emphasizes the balanced evaluation criteria of \texttt{AugCLIP}, which previous metrics fail to abide by. Segment Consistency (SC) excels in MagicBrush since the dataset requires editing a small local area of the image. 
\begin{table}[t]
    \centering
    \caption{
    \textbf{Ground Truth Selection Test.} Accuracy, $\textbf{Acc}_\text{both}$ ($\uparrow$), of assigning higher scores to ground truth images over excessively preserved and modified images.
    }
    \label{tab:exp_main_gt}
    \resizebox{0.9\linewidth}{!}{
        \begin{tabular}{
            @{}
            L{1.5cm}
            M{0.6cm}
            M{0.6cm}
            @{\hskip 0.3in}
            M{2cm}
            M{2cm}
            @{}
            }
        \toprule
        \textbf{Metrics} & Presv. & Modif. &\textbf{TEdBench} & \textbf{MagicBrush} \\
        
        \midrule[0.8pt] 
        \textbf{L2} & \gcmark & \rxmark & 0.310 & 0.002 \\
        \textbf{LPIPS} & \gcmark & \rxmark & 0.090 & 0.000 \\
        \textbf{DINO} & \gcmark & \rxmark & 0.280 & 0.008 \\
        \textbf{SC} & \gcmark & \rxmark & \underline{0.420} & \textbf{0.936} \\
        \textbf{CLIP-I} & \gcmark & \rxmark  & 0.281 & 0.810 \\
        \midrule[0.3pt]
        
        \textbf{CLIP-T} & \rxmark & \gcmark & 0.312 & 0.260 \\
        \textbf{CLIP$_\text{dir}$} & \gtriangle & \gcmark & {0.350} & {0.601} \\
        \midrule[0.3pt]
        \large{\texttt{AugCLIP}} & \gcmark & \gcmark & \textbf{0.570} & \underline{0.889} \\
        \bottomrule
        \end{tabular}
    }
\end{table}

\section{Analysis}\label{sec:anal}

We use \score{} for CelebA, EditVal, and Dreambooth, and \acc{} for TEdBench and MagicBrush. We compare our metric, \texttt{AugCLIP} with \clipdir{}, which is the only metric that considers both preservation and modification.

\paragraph{\texttt{AugCLIP} generalizes to diverse editing scenarios.}
Text-guided image editing encompasses a wide range of tasks such as style editing, object replacement, local editing, partial alteration, and personalized generation. \texttt{AugCLIP} excels in the highly complex scenarios of local editing as conducted with the MagicBrush dataset. Moreover, \texttt{AugCLIP} highly aligns with human judgment in the personalized generation case of DreamBooth, which has been underexplored by previous metrics. We report how well \texttt{AugCLIP} aligns with human judgment with various editing cases provided in the EditVal dataset. \black{Across all eight scenarios, except for the texture modification task, \texttt{AugCLIP} outperforms CLIP$_\text{dir}$ as shown in \cref{tab:comparison}.}

\begin{table}[ht]
    \centering
    \caption{
    \textbf{Human Alignment Score $\vs_\text{2AFC}$ on Various Text-guided Image Editing Scenarios.}}
    \label{tab:comparison}
    \resizebox{\linewidth}{!}{
    \begin{tabular}{@{}lcccc@{}}
        \toprule
        & \textbf{Pos. Add} & \textbf{Obj. repl.} & \textbf{Alter Parts} & \textbf{Background} \\ \midrule
        \textbf{CLIP$_\text{dir}$} & 0.667 & 0.688 & 0.730 & 0.5  \\
        \textbf{\texttt{AugCLIP}} & \textbf{1.0} & \textbf{0.75} & \textbf{0.838} & \textbf{1.0}\\ \bottomrule 
         & \textbf{Texture} & \textbf{Color} & \textbf{Action} & \textbf{Style} \\ \midrule
          \textbf{CLIP$_\text{dir}$} & \textbf{0.806} & \textbf{1.0} & \textbf{1.0} & 0.529 \\
          \textbf{\texttt{AugCLIP}}  & 0.742 & \textbf{1.0} & \textbf{1.0} & \textbf{0.647} \\ \bottomrule
    \end{tabular}
    }
    
\end{table}

\paragraph{Design of Weighting}

To support the rationale behind weighting, we design an experiment\footnote{Two attributes are compared in terms of relative importance in the same context. A total of 10 participants responded in 2AFC format on 40 editing scenarios from the DreamBooth dataset.} that measures \textbf{pearson correlation with human evaluation on attribute importance}, reported in \cref{tab:rebuttal3}. Compared to the naïve weighting strategy mimicking \clipdir{} by measuring cosine similarity between the attribute and the text direction $\Delta T$ (Eq. \textcolor{cornellred}{1}), \texttt{AugCLIP} weighting (Eq. \textcolor{cornellred}{6}) demonstrates impressive alignment with human evaluations.

\begin{table}[h!]
    \centering
    \caption{\footnotesize{Pearson correlation between weighting strategies and human evaluations on attribute importance.}}
    \label{tab:rebuttal3}
    \resizebox{0.8\linewidth}{!}{%
    \begin{tabular}{@{}ccc@{}}
    \toprule
     \multicolumn{1}{c}{Attribute Importance} & \small{Source} & \small{Target} \\
     \midrule
     Naïve Weighting w/ $\Delta T$ & 0.394 & 0.637 \\
     \texttt{AugCLIP} Weighting (Eq. \textcolor{cornellred}{6}) & \textbf{0.794} & \textbf{0.883} \\
    \bottomrule
    \end{tabular}
    }
    \vspace{-10pt}
\end{table}
\paragraph{Augmenting CLIP$_\text{dir}$ cannot provide reliable evaluation result.}
We demonstrate the impact of augmenting CLIP$_\text{dir}$ with attributes, extracted in \cref{subsec:extract_attr}. Specifically, we replace the source and target text embeddings with the average of their respective attribute embeddings, $\mathcal D_S$ and $\mathcal D_T$. As shown in \cref{tab:exp_ablation_clips}, this straightforward augmentation does not guarantee performance gain. \texttt{AugCLIP} consistently outperforms all description-augmented versions of CLIP$_\text{dir}$. This improvement is due to \texttt{AugCLIP}'s approach, which derives the ideal representation by a minimum modification vector. This is fundamentally different from the naïve methodology of CLIP$_\text{dir}$, which relies on a simple subtraction of the source from the target.

\begin{table}[ht]
    \centering
    \caption{\textbf{Effect of Augmenting Attributes.}}
        \label{tab:exp_ablation_clips}
        \resizebox{\linewidth}{!}{%
        \begin{tabular}{lccccc}
        \toprule
        & CelebA & EditVal & DreamBooth & TEdBench & MagicBrush \\ 
       
        \midrule[1.2pt]
        \arrayrulecolor{gray}
        \textbf{CLIP}$_{\text{dir}}$ & 0.673 & 0.697 & 0.357 & 0.350 & \underline{0.601} \\
        \cmidrule(lr){2-4} \cmidrule(lr){5-6}
        $\quad + \mathcal D_S$& 0.816 & 0.629 & 0.357 & 0.400 & 0.429 \\
        $\quad+ \mathcal D_T$ & \underline{0.819} & 0.708 & \underline{0.536} & 0.420 & 0.533 \\
        $\quad+ \mathcal D_S \cup \mathcal D_T$ & 0.816 & 0.607 & \underline{0.536} & \underline{0.440} & 0.402 \\
        \midrule
        \large{\texttt{AugCLIP}} & \textbf{0.883} & \textbf{0.831} & \textbf{0.857} & \textbf{0.570} & \textbf{0.889} \\
        \bottomrule
    \end{tabular}
    }
\end{table}

\paragraph{Weighting Strategy for Hyperplane}
We demonstrate the effectiveness of our weighting strategy, described in \cref{eq:weight} in \cref{tab:exp_ablation_weighting}. Our weighting strategy enables \texttt{AugCLIP} to prioritize key features central to preservation and modification while estimating the representation of an ideally edited image, resulting in improvement in human alignment (\score{}) and balancing preservation and modification (\acc{}). 
\begin{table}[ht]
    \caption{\textbf{Effect of Weighting Strategy.}}
        \label{tab:exp_ablation_weighting}
        \resizebox{\linewidth}{!}{%
        \begin{tabular}{lccccc}
        \toprule
        & CelebA & EditVal & DreamBooth & TEdBench & MagicBrush \\
        \midrule
        Unweighted & 0.849 & 0.787 & 0.786 & 0.400 & 0.830 \\
        Weighted   & \textbf{0.883} & \textbf{0.831} & \textbf{0.857} & \textbf{0.570} & \textbf{0.889} \\
        \bottomrule
        \end{tabular}
        }
\end{table}

\paragraph{Length and Number of Attributes}

In \cref{tab:exp_ablation_desc}, we compare short (under 5 words) and long (over 5 words), as well as cases with a fixed number (10, 20, or 30) of attributes. We observe that using short attributes tends to outperform using long ones, likely because they focus on a single visual aspect, avoiding the potential entanglement of multiple aspects in a single attribute. Moreover, the number of attributes impacts the performance depending on the scene complexity of the benchmark dataset, as complex scenarios require a higher number of attributes to fully capture the scene’s details. Therefore, allowing flexibility in the number of attributes, rather than imposing a strict limit, yields the best overall performance across all configurations.

\begin{table}[ht]
    \caption{\textbf{Effect of Length and Number of Attributes.}}
    \label{tab:exp_ablation_desc}
    \resizebox{\linewidth}{!}{%
    \begin{tabular}{llccccc}
    \toprule
    Length & Number & CelebA & EditVal & DreamBooth & TEdBench & MagicBrush \\
    \midrule
    short & 10 & \underline{0.870} & 0.719 & \textbf{0.857} & \underline{0.540} & \textbf{0.889} \\
    short & 20 & 0.829 & \underline{0.809} & \underline{0.821} & \underline{0.540} & \underline{0.868} \\
    short & 30 & 0.829 & 0.764 & 0.714 & \textbf{0.570} & 0.863 \\
    long  & 30 & 0.843 & 0.697 & 0.750 & 0.530 & 0.862 \\
    \midrule
    short & - & \textbf{0.883} & \textbf{0.831} & \textbf{0.857} & \textbf{0.570} & \textbf{0.889} \\
    \bottomrule
    \end{tabular}
    }
\end{table}
\paragraph{Equivariance of \texttt{AugCLIP} under Description Randomness} Using GPT-4V for description extraction introduces slight randomness across different seeds. To demonstrate near-equivalence quantitatively, we designed a randomness test on the attribute extraction process using the TEdBench dataset, varying five different seeds (examples in Appendix A. \textcolor{cornellred}{2}). \texttt{ROUGE-L} score among attributes extracted by different random seeds is 0.7590 (1 indicates maximum similarity), and the variance of \texttt{AugCLIP} in terms of human alignment is negligible, with a value of 0.0197. We also demonstrate in the Appendix that extracted attributes with different seeds do not vary. 

We provide analysis on the modification vector $\mathbf v$ and choice of classifier function, discussion on limitations, description prompting, and extensive qualitative results in the Appendix.

\section*{Limitations and Discussions}
Our method, \texttt{AugCLIP}, adds additional computation time compared to \clipdir{}. It requires extracting attributes via MLLM, and then fitting a hyperplane to distinguish source and target attributes. This process takes around 12.3 seconds for attribute generation and 0.15 seconds for score computation. This includes the non-reducible CLIP loading time (3.61 sec), which is required by other CLIP-based metrics, and the maximum attribute extraction time (8.62 sec). We will provide extracted attributes for all benchmark datasets to reduce this overhead, facilitating the evaluation process for future usage.

\section*{Conclusion}
We present \texttt{AugCLIP}, a novel evaluation metric for text-guided image editing that balances source image preservation and target text modification. By leveraging a multi-modal language model to extract visual attributes and finding a separating hyperplane, \texttt{AugCLIP} estimates a representation of an ideal edited image that closely matches human preferences. Experiments on five benchmark datasets show that \texttt{AugCLIP} outperforms existing metrics, especially in challenging tasks, offering a more reliable tool for evaluating edits while preserving core attributes.

\clearpage
\section*{Acknowledgements}
This work was supported by the National Research Foundation of Korea (NRF) grants (RS-2023-00209060 A Study on Optimization and Network Interpretation Method for Large-Scale Machine Learning), Institute for Information \& communications Technology Planning \& Evaluation(IITP) grants (No.2022-0-00713 Meta-learning applicable to real-world problems, RS-2019-II190075 Artificial Intelligence Graduate School Program(KAIST)) funded by the Korea government(MSIT).

{
    \small
    \bibliographystyle{ieeenat_fullname}
    \bibliography{main}
}

\onecolumn
\newcommand{\augclip}{\texttt{AugCLIP}}
\newcommand{\clip}[1]{\ensuremath{\text{CLIP}\left(#1\right)}}

\newcommand{\accm}{$\textbf{Acc}_\text{modif}$}
\newcommand{\accp}{$\textbf{Acc}_\text{presv}$}
\clearpage
\appendix

\part{Appendix} 
\parttoc 

\clearpage
\section{Discussions}

\subsection{Intuitive Difference Between \texttt{AugCLIP} and Directional CLIP Similarity}\label{sec:intuition}
\begin{figure}[h]
    \centering
    \includegraphics[width=\linewidth]{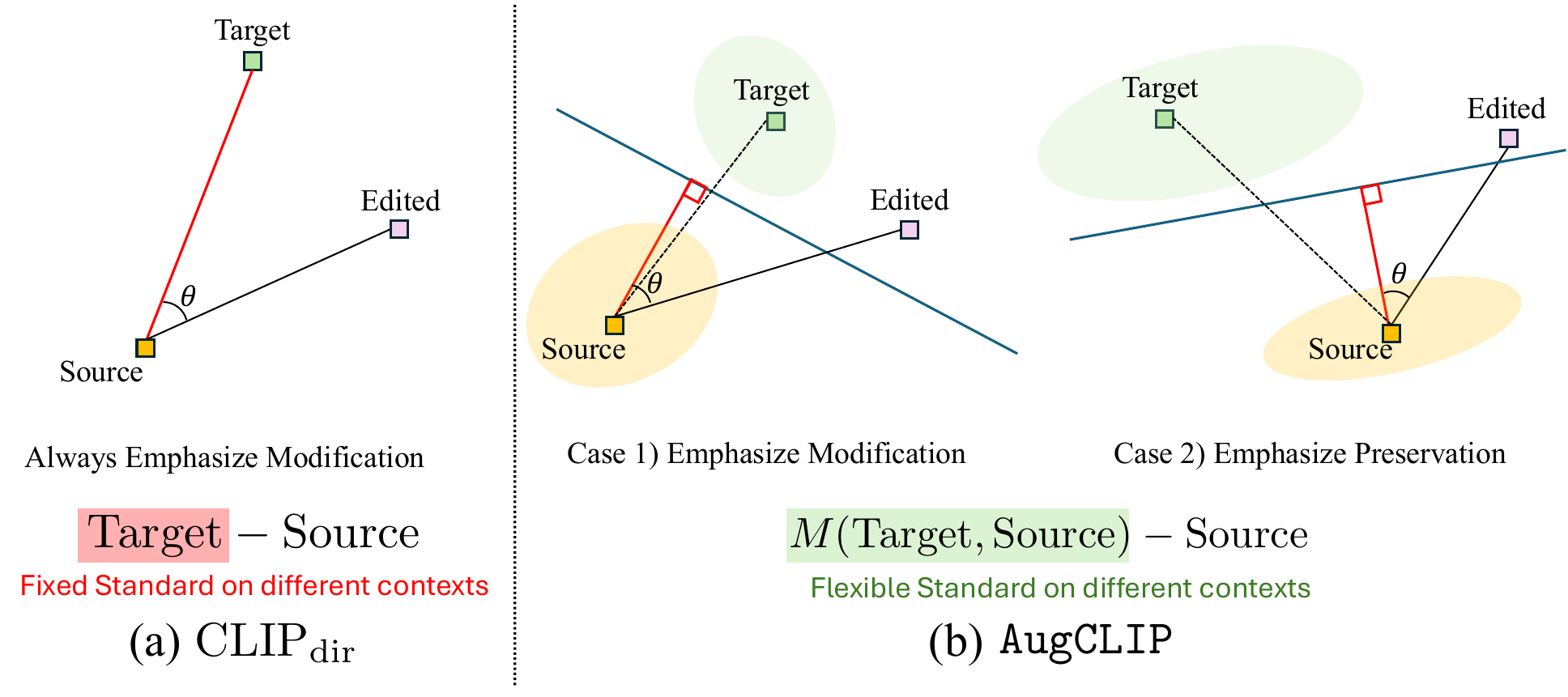}
    \caption{\textbf{Difference between \clipdir{} and \augclip{}.} The red line indicates the evaluation standard and the black line indicates the change in image from source to target. Both \clipdir{} and \augclip{} measure the quality of the edited image according to the corresponding red lines. In \textbf{(b)}, the green and yellow circles indicate the distribution of target and source attributes, respectively, and dotted black lines indicate the evaluation standard of \clipdir{}.}
    \label{fig:intuitive_diff}
\end{figure}

The major difference between \clipdir{} and \augclip{} stems from the flexibility of the evaluation standard. Unlike directional CLIP similarity that relies on the fixed standard of `Target text - Source text' as shown in  \cref{fig:intuitive_diff}\textcolor{cornellred}{(a)}, our metric \augclip{} estimates the contextual difference between the source and the target to flexibly adjust the evaluation standard as `$M(\text{Target, Source}) - \text{Source}$'. More specifically, \cref{fig:intuitive_diff}\textcolor{cornellred}{(b)} shows flexibility of \augclip{}. On the left side, the direction of the evaluation standard (red line) is close to the direction of `Target - Source' as \clipdir{} does. On the right side, the direction of the evaluation standard (red line) does not align with `Target - Source', indicating that the evaluation standard inclines toward preserving the source image, rather than modifying the image into the semantic space of the target text.

In real-world evaluation cases, this is an important property since some tasks might require a small amount of editing that transforms only part of the image whereas some tasks focus on editing the whole image with large modifications. Evaluation metrics should be flexibly applicable to all such cases, deciding the evaluation standard to focus on preservation or modification on a case-by-case basis. However, regardless of the editing context requiring different modification or preservation levels, existing metrics blindly apply the same standard that overly focuses on preservation or modification. 

\subsection{Randomness of Descriptions}
\begin{figure}[htb]
    \centering
    \begin{subfigure}{\linewidth}
        \centering
        \includegraphics[width=\hsize,trim=0 0 0 3cm, clip]{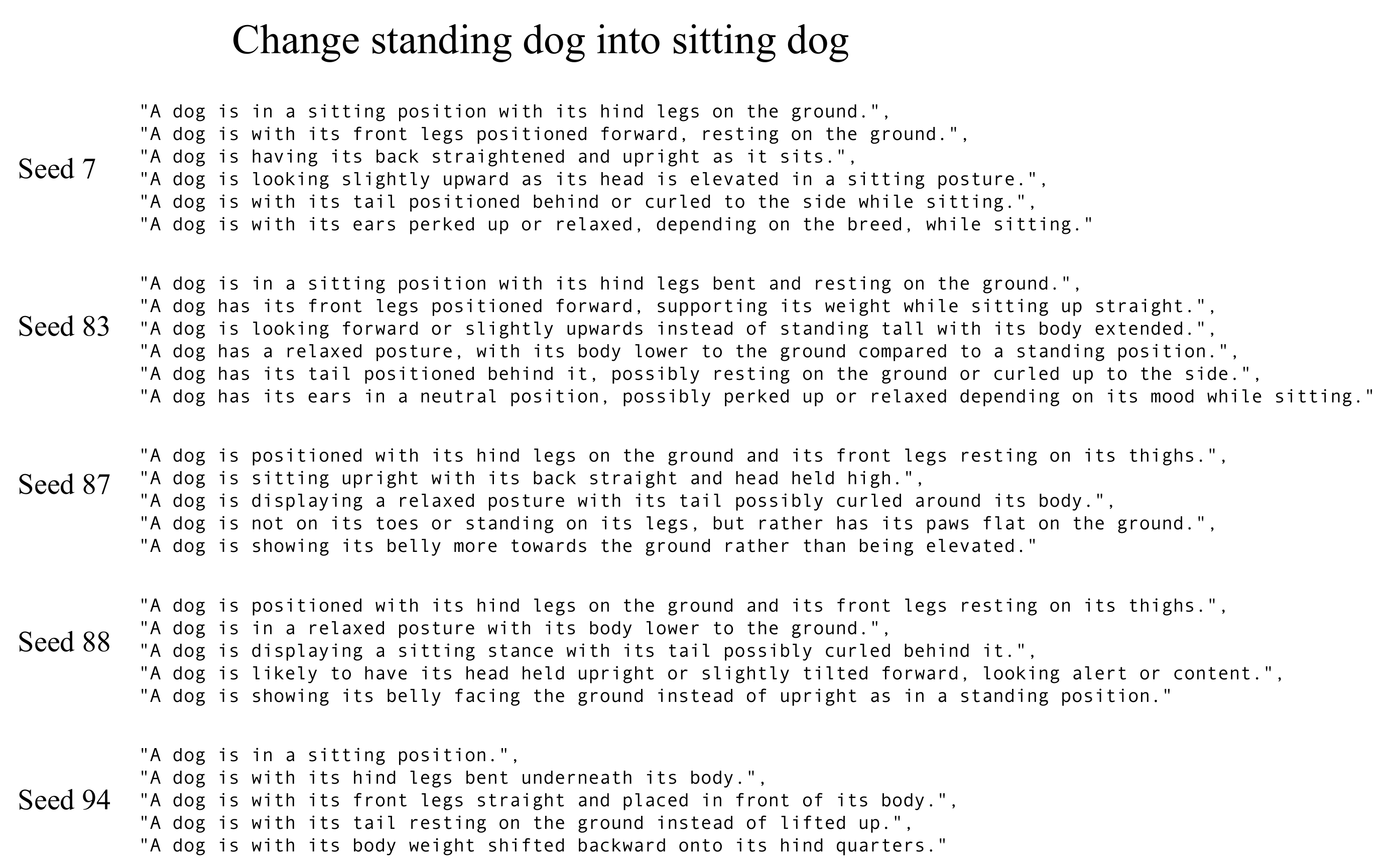}
        \caption{Target text: Change a standing dog into a sitting dog.}
        \label{fig:sub1}
    \end{subfigure}%
    \hfill
    \begin{subfigure}{\linewidth}
        \centering
        \includegraphics[width=\hsize,trim=0 0 0 3cm, clip]{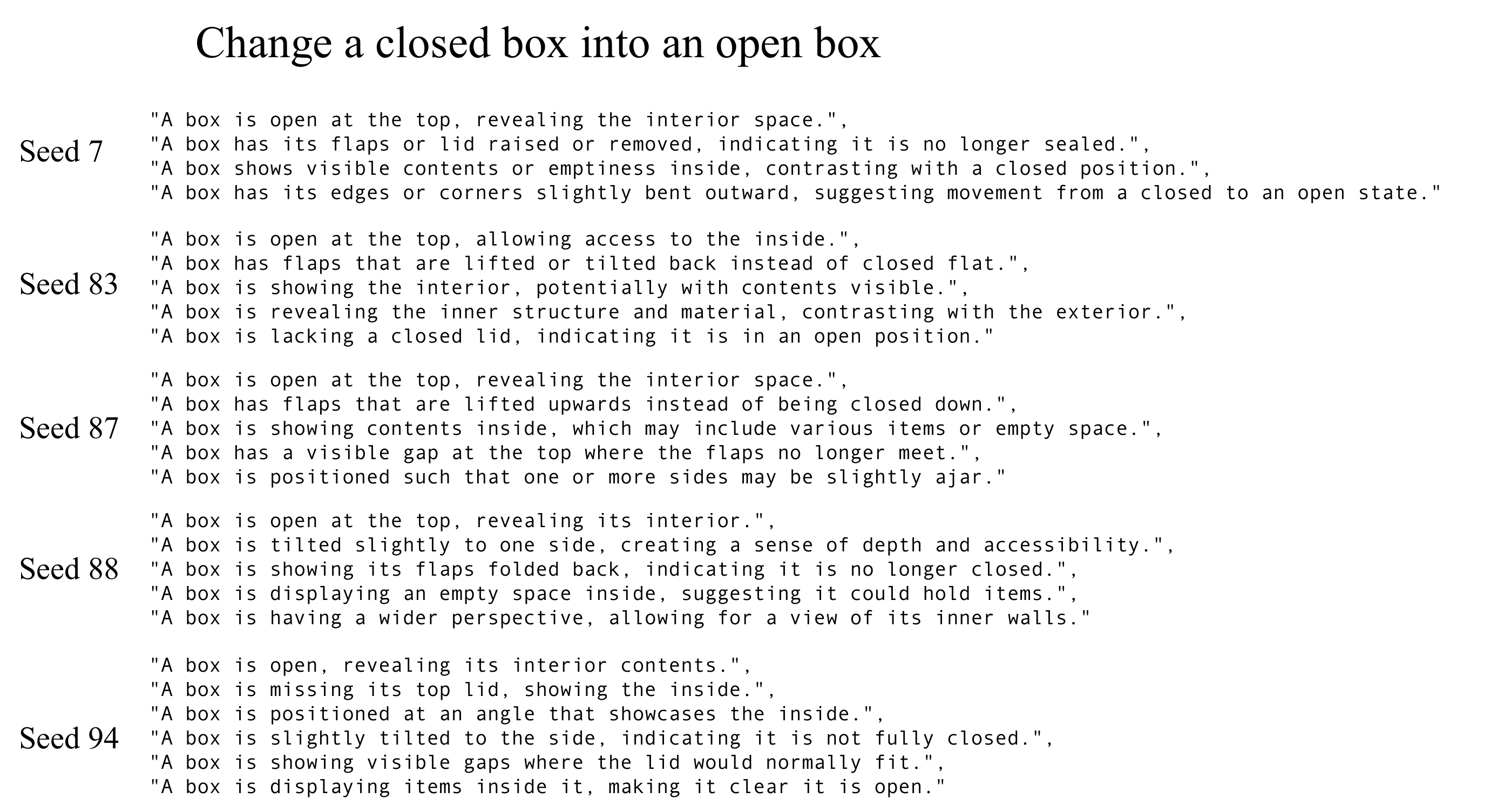}
        \caption{Target text: Change a closed box into an open box.}
        \label{fig:sub2}
    \end{subfigure}
    \caption{\textbf{Target Descriptions Generated with Five Random Seeds.}}
    \label{fig:random_descs1}
\end{figure}

\begin{figure}[htb]
    \centering
    \begin{subfigure}{\linewidth}
        \centering
        \includegraphics[width=\linewidth,trim=0 0 0 3cm, clip]{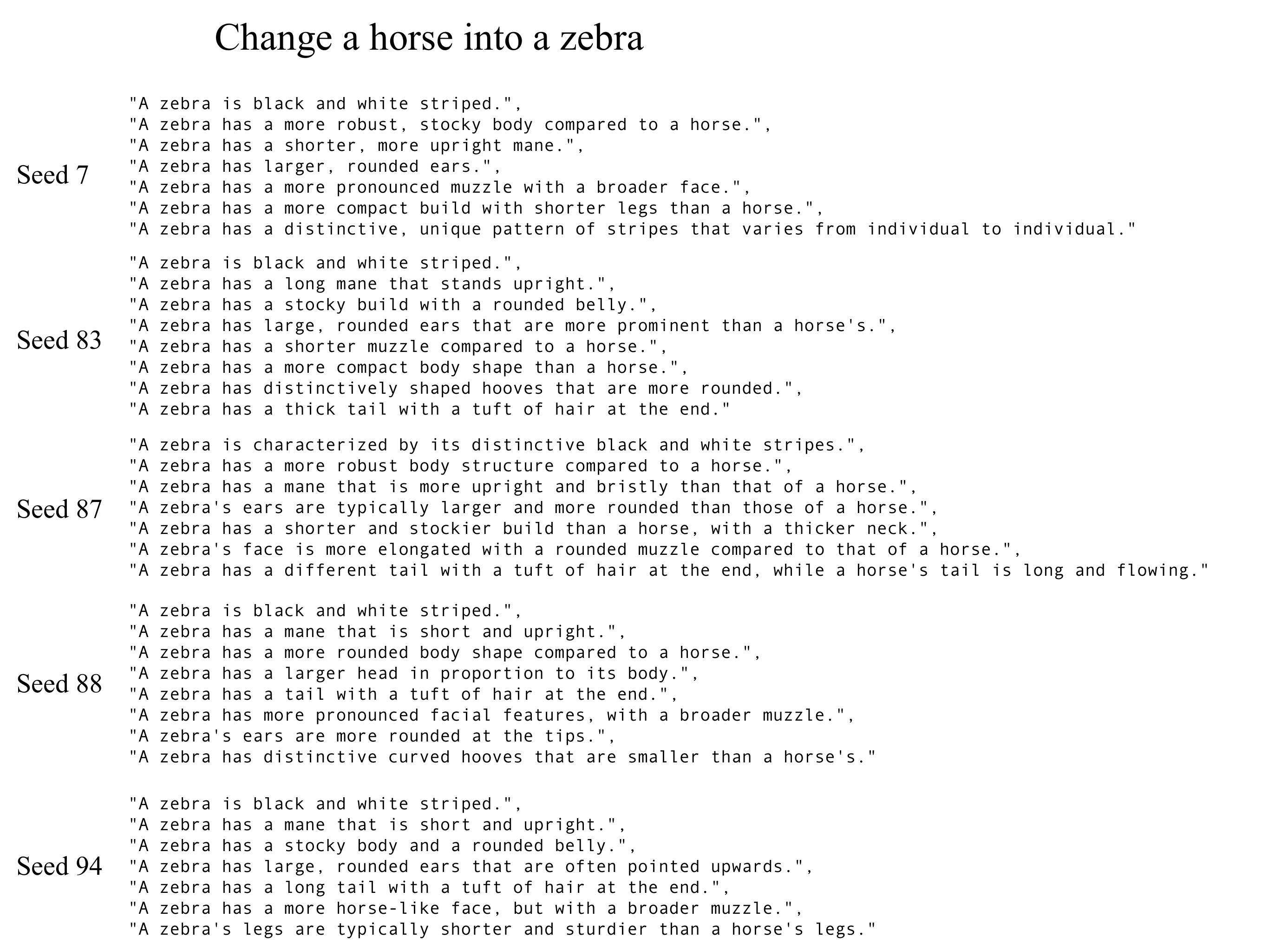}
        \caption{Target text: Change a horse into a zebra.}
        \label{fig:sub3}
    \end{subfigure}
    \hfill
    \begin{subfigure}{\linewidth}
        \raggedright
        \includegraphics[width=0.75\linewidth,trim=0 0 0 3cm, clip]{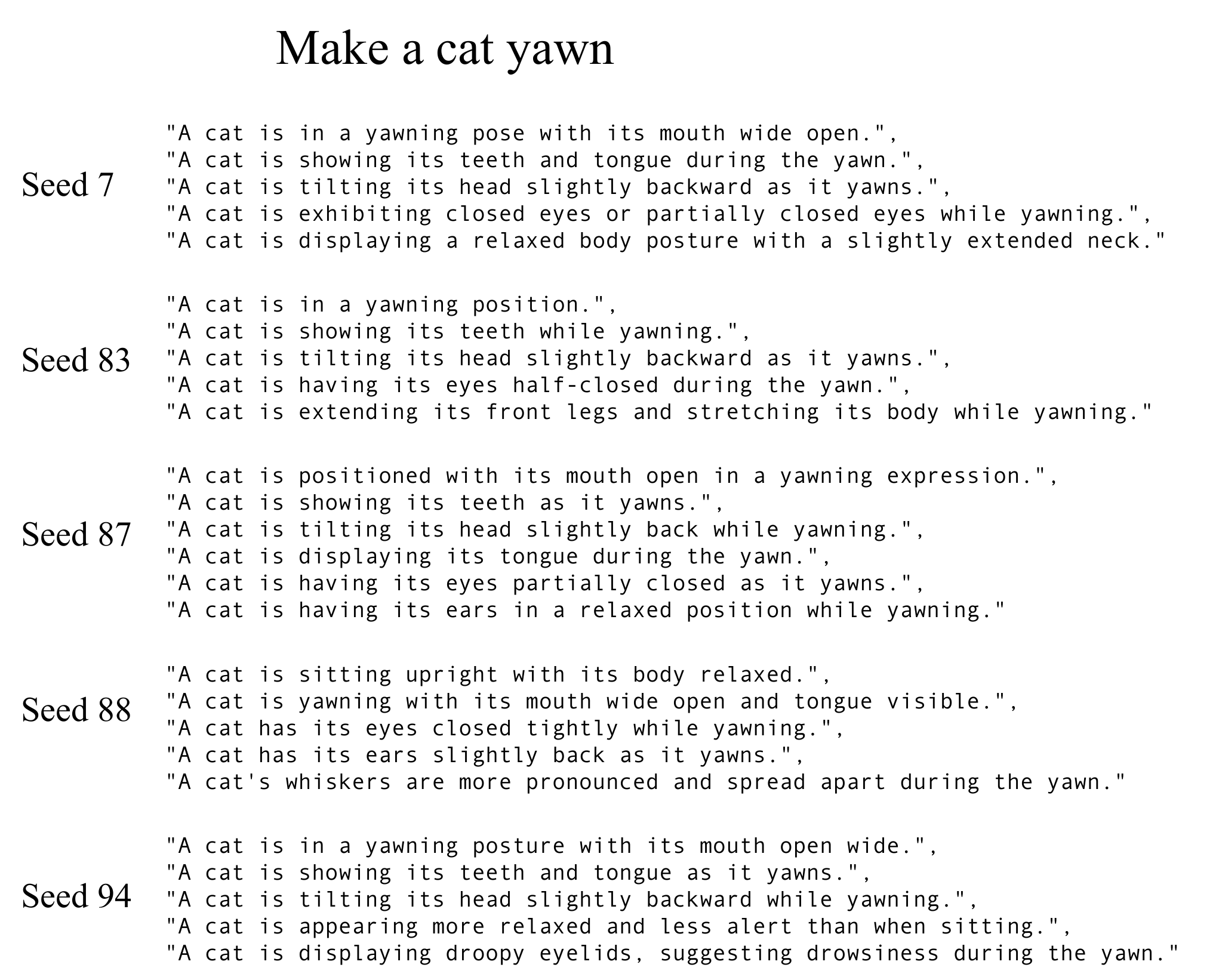}
        \caption{Target text: Make a cat yawn.}
        \label{fig:sub4}
    \end{subfigure}
    \caption{\textbf{Target Descriptions Generated with Five Random Seeds.}}
    \label{fig:random_descs2}
\end{figure}

As discussed in \cref{sec:anal}, the randomness of extracted descriptions does not impact the alignment with human judgment (merely $0.0197$ in variance). In addition to this observation, we provide description samples for the target text, varying the seed over five configurations in \cref{fig:random_descs1}, \ref{fig:random_descs2}. The examples show that generated descriptions have overlapping semantics across different seeds. For example, \cref{fig:sub3} across five seeds describe black and white stripes, with small textual differences but identical in terms of semantics. This is observed across all four examples of the target text, proving that randomness in the description extraction process does not create semantically distinct samples, thus the evaluation results of \augclip{} are robust. the difference across random seeds is almost negligible since source descriptions are directly extracted from the source image caption.

\clearpage

\subsection{Optimization Objective}
In \cref{tab:exp_ablation_hyperplane}, we test three variants of optimization objective for deriving the classifier function $g(x) = \mathbf w ^T x + b$. Since source and target descriptions encoded into CLIP are separable by a simple linear function, we set the hyperplane as $\mathbf w ^T x + b = 0$. Moreover, in optimizing the parameters $\mathbf w$ and $b$, we select the hinge loss objective with L2 regularization, namely SVM objective. 

\paragraph{Reason for Choosing SVM Objective.} We have emonstrate in \cref{tab:exp_ablation_hyperplane}, that this SVM objective is the best option. Moreover, the reasons for choosing this objective are threefold. First, the SVM objective is compatible with various source and target cases. Since SVM allows for misclassification to some extent with the slack variable, it extends to editing cases where the source image and target text do not show distinct discrepancies. Second, support vectors maximize the margin, which is defined as the distance between the hyperplane and the closest support vectors on either source and target descriptions. Third, the influence of data points far from the margin is reduced, since only the closest points (support vectors) determine the decision boundary. This helps avoid bias caused by outlying data points, ensuring the hyperplane is not skewed toward either the preservation or modification side.
\paragraph{Comparing Optimization Objectives.} In order to test if choosing other hyperplane optimization objective impacts the level of alignment with human judgment, we compare latent discriminant analysis (LDA), logistic regression, and linear SVM objective in finding a separating hyperplane. More specifically, the objective functions are as follows,

\begin{align*}
\textsc{LDA} &: \frac{
\text{det}(N_S (\mu_S - \mu)(\mu_S - \mu)^T + N_T (\mu_T - \mu)(\mu_T - \mu)^T)
}
{
\text{det}(\sum_{\vs_i \in \mathcal D_S} (\vs_i - \mu_S)(\vs_i - \mu_S)^T + \sum_{\vt_j \in \mathcal D_T} (\vt_j - \mu_T)(\vt_j - \mu_T)^T)
} \\
\textsc{Logistic} &: - \frac{1}{N} \sum_{i=1}^{N} \left[ y_i \log(
\sigmoid(g(x_i))) + (1 - y_i) \log(1 - \sigmoid(g(x_i))) \right] \\
\textsc{SVM} &: \frac{1}{2} \|\mathbf w \|^2 + C \sum_{i=1}^N \max(0, 1 - y_i \cdot g(x_i))
\end{align*}
where the optimization targets to find $\mathbf w$ and $b$ that minimizes the given objective functions. A total of $N$ pairs of data points $(x, y)$ is employed in the optimization process, where $x$ in these objective functions signifies the CLIP-encoded source or target attributes with the corresponding label $y \in \{-1, 1\}$. 

In \cref{tab:exp_ablation_hyperplane}, we report the human judgment alignment score \score{} and ground truth test accuracy \acc{}. Linear SVM shows the best \score{} and \acc{}, except for CelebA, and the difference between optimization functions do not largely impact the final judgment. Therefore, \augclip{} is robust to the optimization of hyperplane.
\begin{table}[h]
    \centering
    \caption{\textbf{Comparison on difference optimization function for the classifier.}}
        \label{tab:exp_ablation_hyperplane}
        \resizebox{0.8\linewidth}{!}{%
        \begin{tabular}{lcccccc}
        \toprule
        & CelebA & EditVal & DreamBooth & TEdBench & MagicBrush & \makecell{Average Misc.}\\
        \midrule
        LDA & \textbf{0.884} & 0.827 & 0.821 & 0.545 & 0.863 & 3.37\% \\
        Logistic & 0.849 & 0.830 & 0.821 & 0.550 & 0.866 & 1.38\% \\
        Linear SVM   & 0.883 & \textbf{0.831} & \textbf{0.857} & \textbf{0.570} & \textbf{0.889} & \textbf{1.35\%} \\
        \bottomrule
        \end{tabular}
        }
\end{table}

\paragraph{Misclassification Rate.} Our metric \augclip{} first encodes the source and target attributes into CLIP space, denoted as $\mathcal D_S$ and $\mathcal D_T$ respectively. We then employ a linear function $g(x) = \mathbf W^Tx + b$ to estimate a decision boundary that separates the source and target distribution. The average percentage of source and target attributes that are wrongly classified by the hyperplane across five benchmark datasets is reported in \cref{tab:exp_ablation_hyperplane}, denoted as `Average Misc.'. We observe that simple linear decision function $g(x)$ shows a small misclassification rate, $1.35\%$, which signifies its ability to separate the source and target distributions. Specifically, linear SVM achieves the lowest misclassification rate, successfully distinguishing between source and target attributes. Given that the source image and target text may share visual similarities, the extracted source and target attributes cannot always be perfectly separable by a linear hyperplane (\emph{e.g.}, when editing an orange to a tangerine, both the source and edited images share a round shape.). In such cases, these attributes are closely positioned in the embedding space and do not require complete separation. SVM's ability to manage overlapping factors more flexibly allows it to find a more accurate hyperplane, leading to superior performance.

\begin{figure}[ht]
    \centering
    \includegraphics[width=\textwidth]{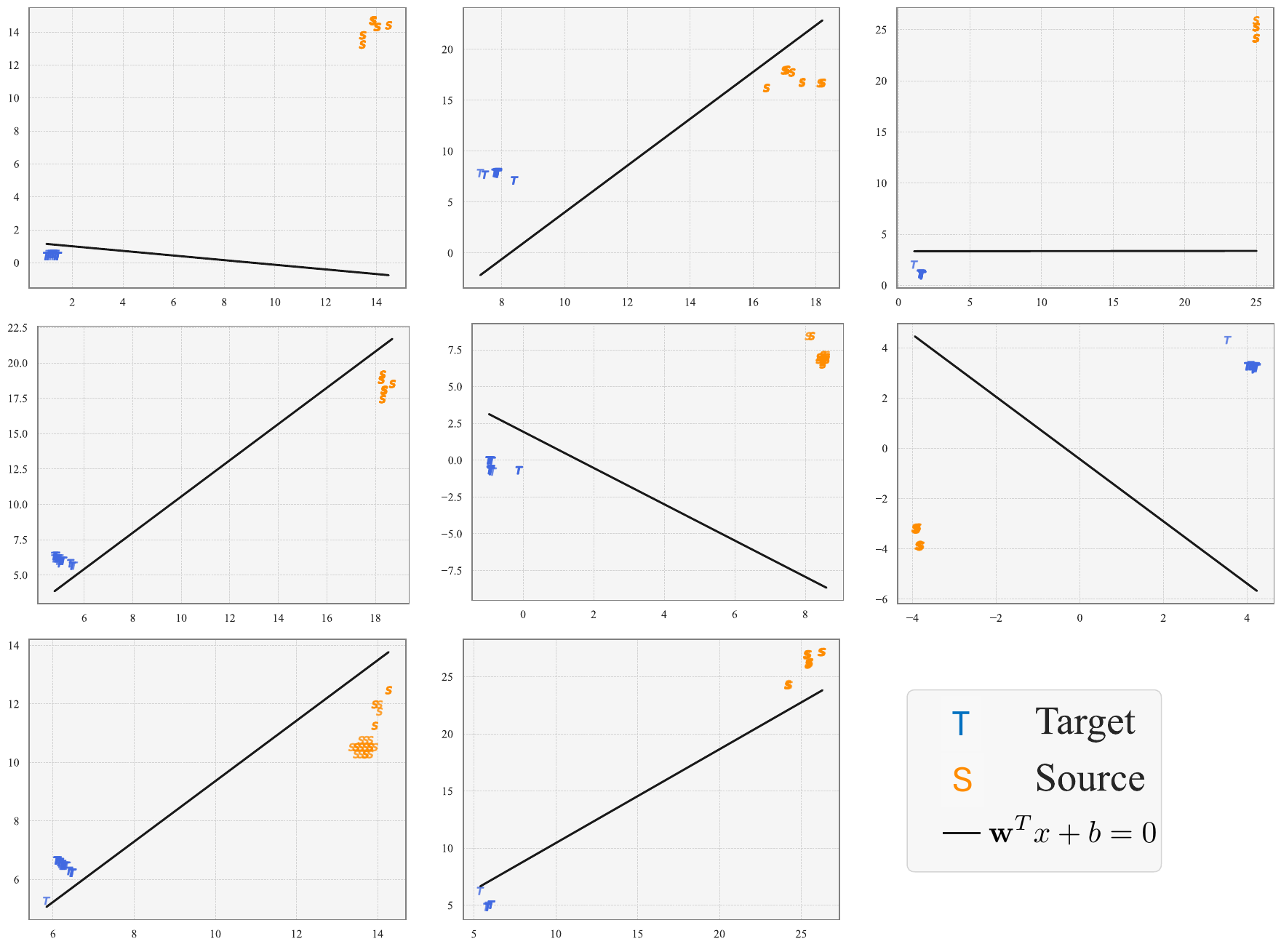}
    \caption{\textbf{Visualization of Source and Target Attributes in 2D Space.} $S$ indicates the source attributes encoded into CLIP space and $T$ indicates the target attributes. The separating hyperplane is a linear function $\mathbf W^Tx + b = 0$.}
    \label{fig:vis-umap}
\end{figure}

\clearpage
\paragraph{UMAP Visualization.} Additionally, we visualize that CLIP features of source and target attributes can be separated by a linear hyperplane in 2D projected space using UMAP \citep{mcinnes2018umap} in \cref{fig:vis-umap}, in which randomly chosen subset of TEdBench samples are plotted. `S' represents source attributes encoded into CLIP, while `T' represents the target attributes. The line signifies the linear hyperplane $g(x) = \mathbf W^Tx + b = 0$ that separates the two classes, which are source and target. In \augclip{}, the linear hyperplane $g(x) = 0$ is $d$-dimensional following the original dimension of CLIP, but to visualize in \cref{fig:vis-umap}, the dimension shrinks into $d=2$ by UMAP fitting.

\clearpage
\subsection{Analyzing the Modification Vector}
The goal of text-guided image editing is to apply a necessary transformation to the source image, preserving the original property as much as possible. Our evaluation process follows this protocol by estimating the modification vector $\mathbf v$ as a minimum modification that makes the source image look like a target text, ensuring that essential source attributes remain unchanged but the resulting edited image resembles the target text. In this section, we qualitatively analyze the effect of $\mathbf v$ on source and target attributes.

\begin{figure}[ht]
    \centering
    \includegraphics[width=\textwidth]{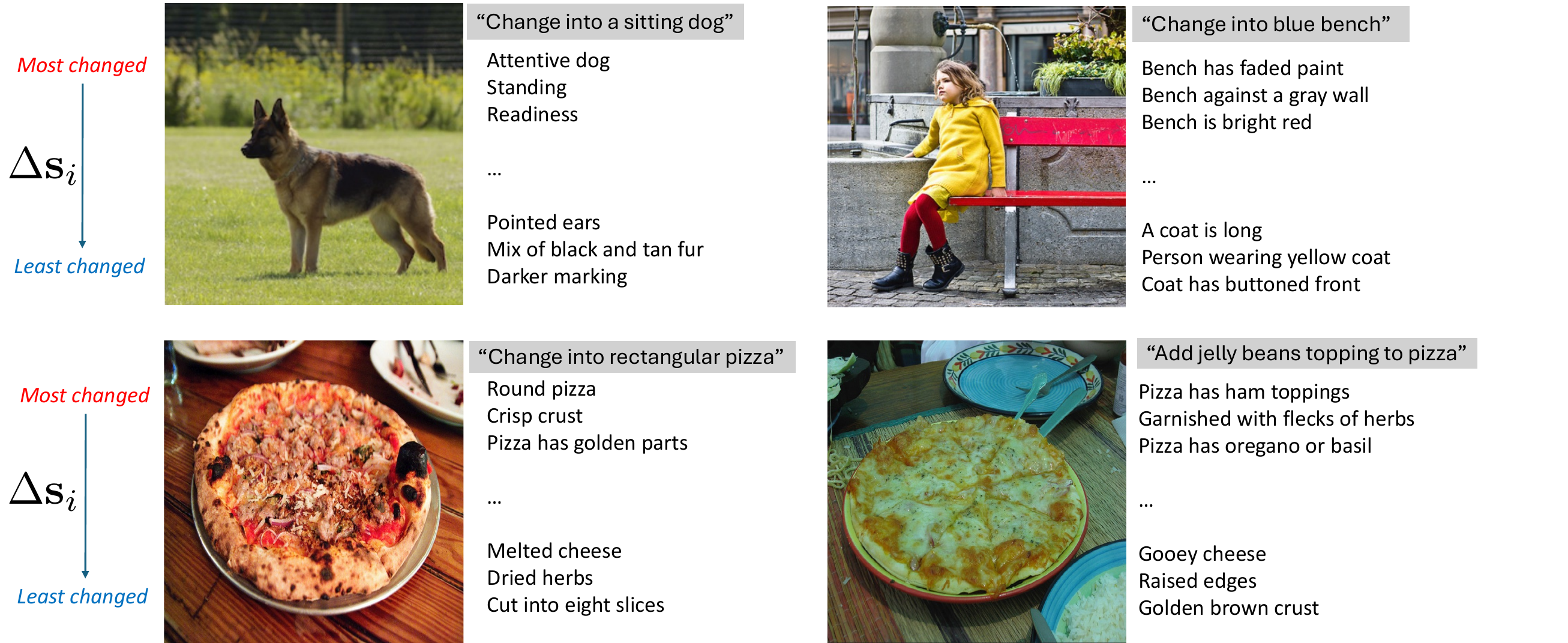}
    \caption{\textbf{Effect of the Modification Vector $\mathbf v$ on Source Attributes.} The source attributes are ranked based on the magnitude of change induced by the modification vector $\mathbf v$. Attributes at the top of the list exhibit the most significant adjustments toward the target text, indicating that these characteristics are considered important source attributes that should be modified in the given editing context. Conversely, attributes lower down the list are determined to remain intact in an ideal editing.}
    \label{fig:analyze_sproj}
\end{figure}

\paragraph{Effect of the Modification Vector $\mathbf v$ on Source Attributes} 
Source attributes that need to be preserved should not be affected by the modification vector $\mathbf v$, while those requiring adjustment according to the target text should be altered. To demonstrate that $\mathbf v$ drives significant changes in the attributes requiring modification while minimally impacting those that should remain intact, we analyze several cases using the TEdBench and EditVal datasets, as shown in \cref{fig:analyze_sproj}. We measure the difference in cosine similarity between each source attribute $\vs_i$ and both the source image and the ideally edited image. Specifically, we calculate the increase in similarity between the ideal edited image  $E(I_\text{src}) + \mathbf v$, and the source image $E(I_\text{src})$ as 
\begin{equation}
\Delta \vs_i = \texttt{cs}\Big(E(I_\text{src}) + \mathbf v, \vs_i\Big) - \texttt{cs}\Big(E(I_\text{src}), \vs_i\Big),
\end{equation}
where $\vs_i$ indicate the CLIP-encoded source attribute in $\mathcal D_S$.
As demonstrated in \cref{fig:analyze_sproj}, source attributes that need to be preserved exhibit only small changes in similarity (small $\Delta \mathbf s_i$), while attributes that need modification (large $\Delta \mathbf s_i$) show a significant change in similarity.

\clearpage
\begin{figure}[ht]
    \centering
    \includegraphics[width=\textwidth]{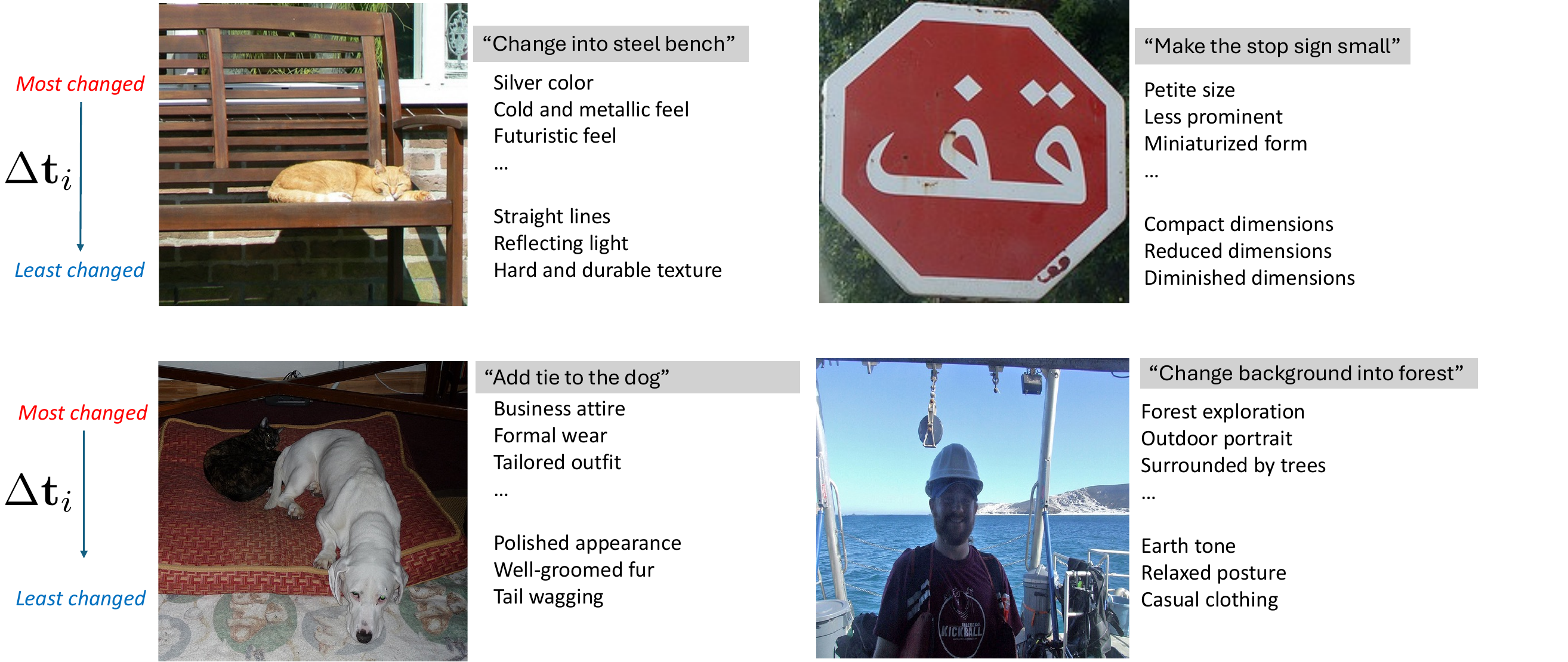}
    \caption{\textbf{Effect of the Modification Vector $\mathbf v$ on Target Attributes.} The target attributes are ranked based on the magnitude of change induced by the modification vector $\mathbf v$. Attributes at the top of the list are determined by \augclip{} standard to be key to modification aspects. Attributes lower on the list exhibit smaller adjustments, which are deemed less important by \augclip{} standard.}
    \label{app:v_trg_attr}
\end{figure}

\paragraph{Effect of the Modification Vector $\mathbf v$ on Target Attributes}
We demonstrate how target attributes are affected by the modification vector $\mathbf v$. To demonstrate that $\mathbf v$ causes significant changes in the attributes that are central to make the image resemble the target text while having minimal impact on the rather peripheral attributes, we analyze several cases using the EditVal dataset, as shown in \cref{app:v_trg_attr}. The difference in cosine similarity between the source image and the ideally edited image is measured for each target attribute $\vt$. Similar to the previous paragraph, we compare the increase of target attribute $\vt_i$ in the ideal edited image,  $E(I_\text{src}) + \mathbf v$, compared to the source image $E(I_\text{src})$ as
\begin{equation}
\Delta \vt_i = \texttt{cs}\Big(E(I_\text{src}) + \mathbf v, \vt_i\Big) - \texttt{cs}\Big(E(I_\text{src}), \vt_i\Big),
\end{equation}
where $\vt_i$ indicate the CLIP-encoded target attribute in $\mathcal D_T$.
As illustrated in \cref{app:v_trg_attr}, attributes essential for matching the target text display a significant increase in similarity, while secondary attributes experience only minor changes.

\clearpage
\section{Algorithm}
We provide the algorithm of \augclip{} in Python code style in the following block. 

\begin{lstlisting}
# Step 0: Get CLIP features
src_img_feat, tgt_img_feat = CLIP(src_img), CLIP(tgt_img)
src_text_feat, tgt_text_feat = CLIP(src_text), CLIP(tgt_text)
src_desc_feat, tgt_desc_feat = CLIP(src_desc), CLIP(tgt_desc)

# Step 1: Compute importance weighting for each desc
src_dist = [src_img_feat, src_desc_feat]
tgt_dist = [tgt_img_feat, tgt_desc_feat]

src_weight, tgt_weight = compute_weight(src_dist, tgt_dist)
weight = [src_weight, tgt_weight]

# Step 2: Fit the classifier model
X = [src_dist, tgt_dist]
y = [-1] * src_dist.shape[0] + [1] * tgt_dist.shape[0]
svc_classifier.fit(X, y, sample_weight=weight)

# Step 3: Retrieve the hyperplane parameters
w = svc_classifier.coef_  # Hyperplane coefficients
b = svc_classifier.intercept_  # Hyperplane intercept

# Step 4: Compute the modification vector v by calculating the projection of src_img_feat onto the hyperplane
numerator = -(np.dot(w.T, src_img_feat) + b)
denominator = np.linalg.norm(w)**2
v = (numerator / denominator) * w

# Step 5: Calculate the alignment score
score = cosine_similarity(src_img_feat + v, tgt_img_feat)
\end{lstlisting}

\clearpage
\section{Evaluation Details}

\subsection{Assets} \label{subsec:app_assets}

\begin{table}[ht]
\centering
\caption{\textbf{Assets Employed in Our Experiments.} List of pre-trained models, benchmark datasets, and metrics employed in this paper. }
\label{tab:assets}
    \resizebox{\textwidth}{!}{%
        \begin{tabular}{@{}lcll@{}}\toprule
         Category & & Asset & URL \\ \midrule
         \multirow{5}{*}{\textbf{Benchmarks}} 
         & & CelebA~\citep{liu2015faceattributes} & \url{https://mmlab.ie.cuhk.edu.hk/projects/CelebA.html} \\
         & & TedBench~\citep{kawar2022imagic} & \url{https://github.com/imagic-editing/imagic-editing.github.io/tree/main/tedbench} \\
         & & EditVal \citep{basu2023editval} & \url{https://github.com/deep-ml-research/editval_code} \\
         & & DreamBooth \citep{ruiz2023dreambooth} & \url{https://github.com/google/dreambooth} \\
         & & MagicBrush \citep{zhang2024magicbrush} & \url{https://github.com/OSU-NLP-Group/MagicBrush} \\
         \midrule
         
         \multirow{11}{*}{\textbf{Editing Models}} & 
         \multirow{5}{*}{\makecell{TEdBench \\ EditVal \\ MagicBrush}}
         & InstructPix2Pix~\citep{brooks2022instructpix2pix} & \url{https://github.com/timothybrooks/instruct-pix2pix}\\
         & & DiffEdit~\citep{couairon2022diffedit} & \url{https://github.com/Xiang-cd/DiffEdit-stable-diffusion.git} \\
         & & Prompt-to-Prompt~\citep{hertz2022prompt} & \url{https://github.com/google/prompt-to-prompt.git}\\
         & & DDS~\citep{hertz2023delta} & \url{https://github.com/google/prompt-to-prompt/blob/main/DDS_zeroshot.ipynb}\\
         & & Imagic~\citep{kawar2022imagic} & \url{https://github.com/huggingface/diffusers/tree/main/examples/community\#imagic-stable-diffusion}\\
         
         \cmidrule[0.4pt]{2-4}
         & \multirow{4}{*}{CelebA} & DiffusionCLIP~\citep{kim2021diffusionclip} & \url{https://github.com/gwang-kim/DiffusionCLIP}\\
         & & Multi2One~\citep{kim2022learning} & \url{https://github.com/akatigre/multi2one}\\
         & & Asyrp~\citep{kwon2022diffusion} & \url{https://github.com/kwonminki/Asyrp_official}\\
         & & StyleCLIP~\citep{Patashnik_2021_ICCV} & \url{https://github.com/orpatashnik/StyleCLIP}\\
         
         \cmidrule[0.4pt]{2-4}
         & \multirow{3}{*}{DreamBooth} & Custom Diffusion ~\citep{kumari2023multi} & \url{https://github.com/adobe-research/custom-diffusion}\\
         & & BLIP-Diffusion~\citep{li2024blip} & \url{https://github.com/salesforce/LAVIS/tree/main/projects/blip-diffusion}\\
         & & ELITE~\citep{wei2023elite} & \url{https://github.com/csyxwei/ELITE}\\
        \midrule
        \multirow{4}{*}{\textbf{Metrics}} & &
        CLIP \citep{clip} & \url{https://github.com/openai/CLIP} \\
        & &
        LPIPS~\citep{zhang2018unreasonable} & \url{https://pypi.org/project/lpips/} \\
        & & SC~\citep{kim2021diffusionclip} & Implemented by the Authors \\
        \bottomrule
        \end{tabular}
    }
\end{table}

\clearpage
\subsection{Source Description Generation}\label{app:src_prompt}

\newtcolorbox{redbox}[1]{colback=red!5!white,colframe=red!85!black,fonttitle=\bfseries,title=#1,width=0.8\textwidth,left=5pt,right=5pt,boxrule=1pt}
\newtcolorbox{bluebox}[1]{colback=blue!5!white,colframe=blue!85!black,fonttitle=\bfseries,title=#1,width=0.8\textwidth,left=5pt,right=5pt,boxrule=1pt}
\newtcolorbox{greenbox}[1]{colback=green!5!white,colframe=green!75!black,fonttitle=\bfseries,title=#1,width=0.8\textwidth,left=5pt,right=5pt,boxrule=1pt}
As described in \cref{subsec:extract_attr}, we prompt GPT-4V with the source image to first generate a detailed caption. Specifically, the prompt orders to focus on the visual and objective aspects of the image, disregarding any opinions or non-visual descriptions. With the generated caption, we provide an example of another source image caption and its parsed list of visual characteristics. With these examples, GPT-4V parses the source image caption into a Python list of visual characteristics. The example prompting result is shown as follows. 

\begin{center}
\begin{redbox}{User}
    \begin{minipage}[m]{0.3\textwidth} 
        \centering
        \includegraphics[width=0.9\linewidth]{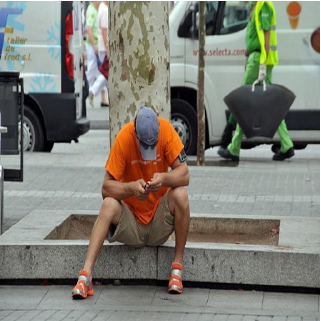} 
    \end{minipage}%
    \hspace{0.02\textwidth} 
    \begin{minipage}[m]{0.6\textwidth} 
        \raggedright
        Describe the image in detail. Do not include opinions or non-visual aspects. 
        Make the descriptions easy and intuitive.
    \end{minipage}
\end{redbox}

\begin{bluebox}{GPT-4V}
    The image shows a person sitting on a low ledge. He is wearing an orange T-shirt and tan shorts. His legs are spread apart, and he appears to be focused on his phone, which he is holding with both hands. He has a gray cap on his head, and his athletic shoes are orange with gray accents. His posture is slightly hunched over as he engages with his device.
\end{bluebox}

\begin{redbox}{User}
You are a helpful text-generation assistant. Given detailed textual descriptions about an image, your goal is to parse into specific visual attributes.\\
\textsc{EXAMPLE INPUT}: ``The image features a large, dark-colored dog standing in a grassy field. The dog appears to be alert and attentive, possibly observing its surroundings. The grass is lush and green, providing a natural backdrop for the dog.'' \\\\
\textsc{EXAMPLE OUTPUT}: [``A dog is large'', ``A dog is dark-colored'', ``A dog is standing'', ``A dog is standing on a grassy field'', ``A dog is alert'', ``A dog is attentive'', ``A dog is observing its surroundings'', ``A grass is lush'', ``A grass is green'', ``A grass is providing a natural backdrop''] \\
\par\noindent\rule{\textwidth}{0.4pt}
Give me a Python list of visual characteristics included in the following sentences: \\
\textsc{INPUT}: ``The image shows a person sitting on a low ledge. He is wearing an orange T-shirt and tan shorts. His legs are spread apart, and he appears to be focused on his phone, which he is holding with both hands. He has a gray cap on his head, and his athletic shoes are orange with gray accents. His posture is slightly hunched over as he engages with his device.''
\end{redbox}

\begin{bluebox}{GPT-4V}
    [``He is sitting on a low ledge'', ``He is wearing an orange t-shirt'', ``He is wearing tan shorts'',
    ``He has his legs spread apart'',
    ``He is focused on his phone'',
    ``He is holding the phone with both hands'',
    ``He has a gray cap on his head'',
    ``He has athletic shoes'',
    ``He has orange shoes with gray accents'',
    ``He is slightly hunched over'']
\end{bluebox}
\end{center}

\subsection{Target Description Generation}\label{app:trg_prompt}
The descriptions for the target text are generated by providing the target text and the source image caption. The prompt instructs the model to avoid repetition of similar visual attributes. The example depicts a case of changing a sitting person into a standing posture. The output is structured into a Python list of descriptions, starting with ``Standing person''. From the target text ``Change a man into Ironman,'' the generated list of attributes includes descriptions of the suit, weapons, and armor. These also depict that Ironman takes heroic poses and appears in intense battles. The example prompting result is shown as follows. 

\begin{center}
\begin{redbox}{User}
You are a helpful text-generation assistant. Given a detailed textual description of an image, your goal is to parse it into specific visual attributes. If the visual attributes with similar meanings are repeated, only use one. For example, ``A dog is large'' and ``A dog is big'' are similar, so only use one.\\\\
\textsc{EXAMPLE INPUT}: In order to make a person into a standing posture, what visual attributes of the
image should be changed? Answer in the format of ``Standing person is ATTRIBUTE''.\\\\
\textsc{EXAMPLE OUTPUT}: [``Standing person has straight legs'', ``Standing person is upright'', ``Standing person
is on their feet'', ``Standing person is tall'']\\
\par\noindent\rule{\textwidth}{0.4pt}
\textsc{INPUT}: In order to make a person into Ironman, what visual attributes of the
image should be changed? Answer in the format of ``Ironman is ATTRIBUTE''.
\end{redbox}

\begin{bluebox}{GPT-4V}
    [``Ironman has powered exoskeleton suit'',
``Ironman has a red and gold color scheme'',
``Ironman has Arc reactor on chest'',
``Ironman wears metallic armor'',
``Ironman wears helmet with facial display'',
``Ironman has repulsor technology on hands'',
``Ironman has jet boots for flight'',
``Ironman has advanced weapon systems'',
``Ironman has a strong and muscular build'',
``Ironman has heroic poses'',
``Ironman appears in intense battles''
]
\end{bluebox}
\end{center}

\clearpage
\subsection{User Study Details} \label{subsec:app_survey}

\begin{figure}[ht]
    \centering
    \includegraphics[width=\linewidth]{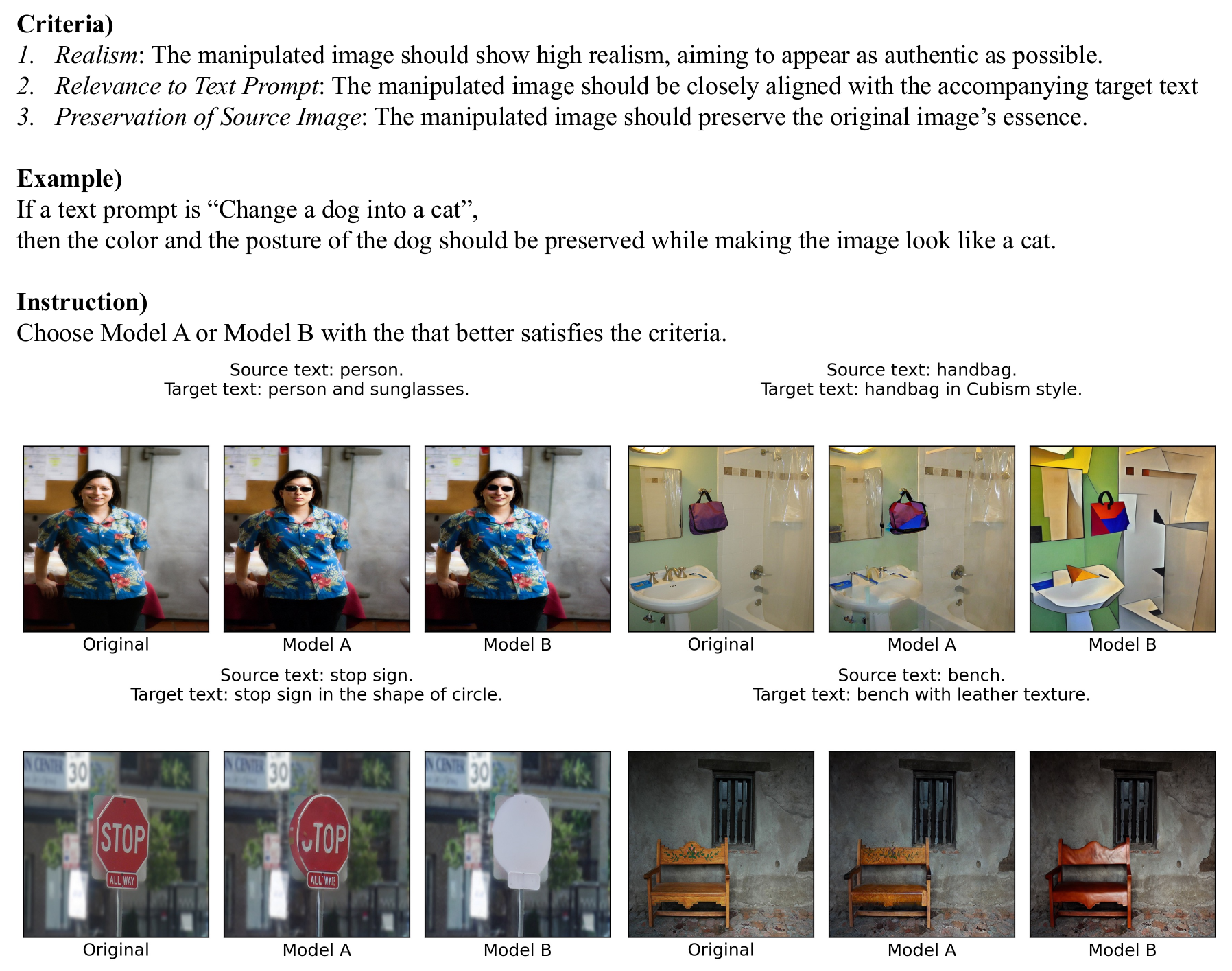}
    \caption{\textbf{User Study Details.} The figure shows the instructions and example questions of our user study.}
    \label{fig:survey}
\end{figure}

As existing text-guided image editing models do not guarantee adequate visual quality, we manually select the images that show sufficient change in the image to conduct a user study. Each participant is provided with a source image, its corresponding target text, and two variants of edited images. Then, human evaluators are instructed to choose the image with better editing quality. As shown in \cref{fig:survey}, clear guidelines are provided to instruct the participants to evaluate the images based on both the preservation of the source image and the modifications toward the target text.

\subsection{Benchmark Datasets}\label{app:benchmark}

\textbf{TEdBench} comprises $100$ pairs of source image and target text. It focuses on specific settings where the source image has a single object at the center, and the corresponding target text only modifies some attributes of that object. 

\textbf{EditVal} contains $648$ image-text pairs that cover $13$ different types of edits, including object addition, object replacement, and size modification. Since it has such complicated editing scenarios, the models that we use for editing could not properly edit the majority of the cases, leaving almost no samples with enough quality for user study. Therefore, we use the subset of EditVal, which encompasses eight editing types that show adequate modification for proper evaluation. 

\textbf{MagicBrush} is a benchmark specifically designed to evaluate sequential editing tasks, where iterative modifications are made to different parts of the source image. 

\textbf{Dreambooth} enables the modification of specific instances within the source image by providing corresponding masks along with image-text pairs; however, since typical editing models do not utilize masks as input, we only consider the image-text pairs in our evaluation. 

\textbf{CelebA} dataset consists of $50$ image-text pairs that guide changes specific to facial attributes. We create target texts by swapping attributes of human faces. 

\clearpage
\section{Additional Results}
\subsection{Combination of Preservation and Modification Centric Metrics}\label{app:comb_PM}
\begin{figure}[ht]
    \centering
    \includegraphics[width=0.8\textwidth]{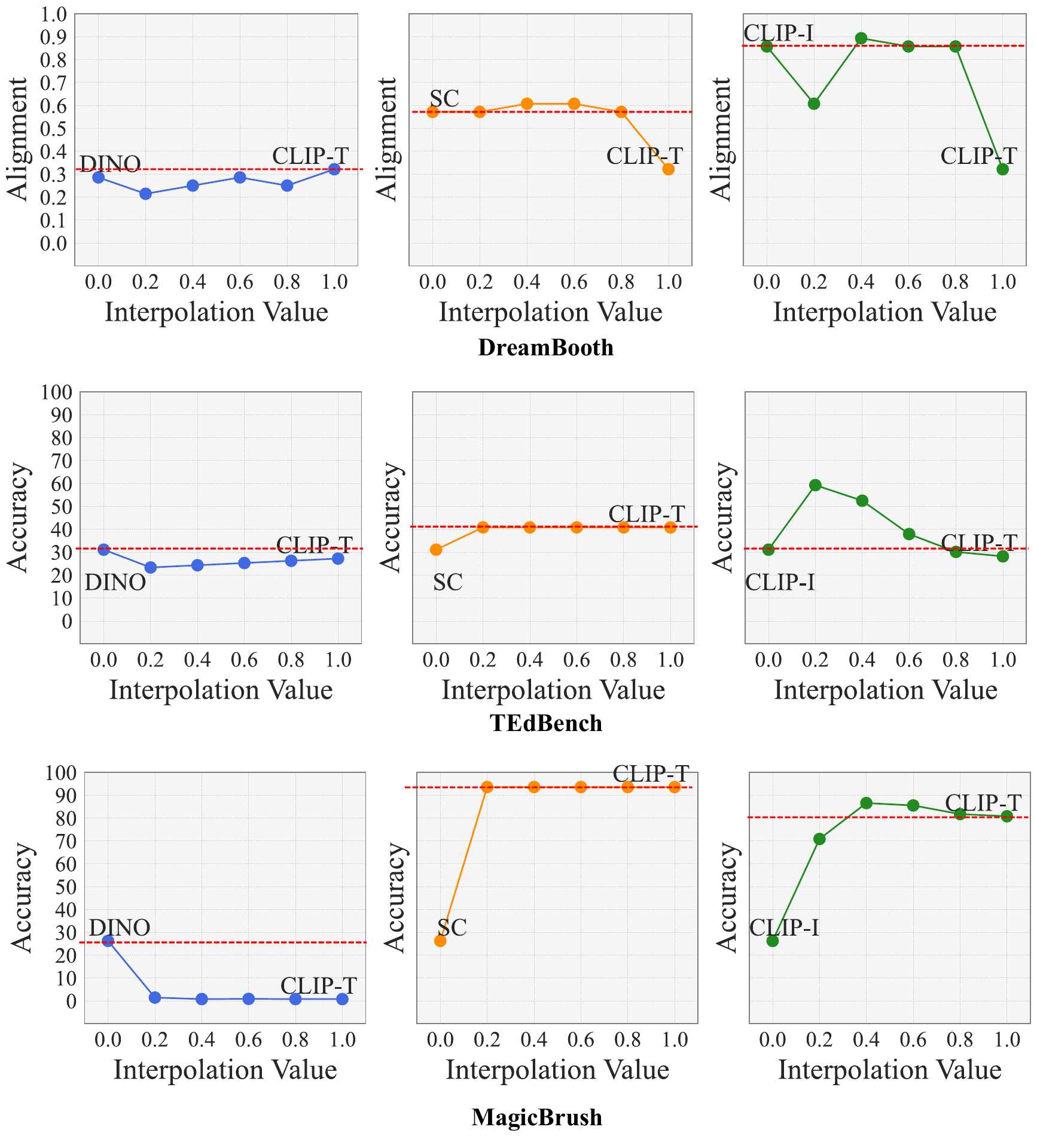}
    \caption{\textbf{Interpolation of Preservation- and Modification-Centric Metrics.}}
    \label{fig:app_interp}
\end{figure}

As discussed in \cref{subsec:comb_prob}, combining modification-centric metric (CLIP-T) with existing preservation-centric metrics (DINO similarity, Segment Consistency, CLIP-I) shows negligible improvement or rather deteriorates in terms of alignment with human judgment and ground truth selection test. Due to the spatial constraint, we have shown two of the datasets, EditVal and CelebA, in the main paper. In \cref{fig:app_interp}, we demonstrate the results of the other three datasets. Notably, the combination of CLIP-I and CLIP-T shows improvement in the TEdBench dataset. Since TEdBench is the dataset of the simplest setting, where source images are highly object-centric and the target text instructs relatively simple modification, a simple strategy of combining these two metrics could be a viable option for evaluation. However, as shown in most editing cases, such a simplistic combination approach fails to show large improvement, underscoring the need for a metric tailored for text-guided image editing.

\subsection{Additional Result on Augmenting Directional CLIP Similarity}

\begin{table}[ht]
    \centering
    \caption{\textbf{Effect of Augmenting Attributes into \clipdir{}.} We use \score{} for CelebA, EditVal, and Dreambooth, and \acc{} for TEdBench and MagicBrush.}
    \label{tab:app_clips_aug}
    \resizebox{0.9\linewidth}{!}{%
    \begin{tabular}{lcccccc}
    \toprule
    &\black{Weighting} & CelebA & EditVal & DreamBooth & TEdBench & MagicBrush \\
    \midrule
    \textbf{\clipdir{}} & & 0.673 & 0.697 & 0.357 & 0.350 &\underline{0.601} \\
    \arrayrulecolor{gray}
    \cmidrule{2-7}
    \multirow{2}{*}{\textbf{ + src attr.}} & \rxmark & 0.816 & 0.629 & 0.357 & 0.400 & 0.429 \\
    & \gcmark & 0.796 & \underline{0.725} & 0.464 & 0.440 & 0.523 \\
    \cmidrule{2-7}
    \multirow{2}{*}{\textbf{ + trg attr.}} & \rxmark & \underline{0.819} & 0.708 & \underline{0.536} & 0.420 & 0.533 \\
    & \gcmark & 0.734 & 0.513 & 0.464 & 0.450 & 0.443 \\
    \cmidrule{2-7}
    \multirow{2}{*}{\textbf{ + src \& trg attr.}} & \rxmark & 0.816 & 0.607 & \underline{0.536} & {0.440} &  0.402 \\
    & \gcmark & 0.636 & 0.600 & \underline{0.536} & \underline{0.500} & 0.407 \\
    \midrule
    \large{\augclip{}} & \gcmark & \textbf{0.883} & \textbf{0.831} & \textbf{0.857} & \textbf{0.570} & \textbf{0.889} \\
    \bottomrule
    \end{tabular}
    }
\end{table}

In \cref{tab:exp_ablation_clips}, we demonstrate that the simple strategy of directly augmenting \clipdir{} with the source and target attributes fails to outperform \augclip{}. Additionally, we show the effect of applying the weighting strategy (\gcmark) of \cref{eq:weight} when aggregating attributes into \clipdir{} in \cref{tab:app_clips_aug}. Note that augmentation of \clipdir{} without weighting (\rxmark) is already reported in \cref{tab:exp_ablation_clips}. 

Formally, for augmenting the source text $T_\text{src}$ with source attributes in $\mathcal D_S$, directional CLIP similarity is redefined as
\begin{equation}
    \texttt{cs}\Big(E(I_\text{edit}) - E(I_\text{src}), E(T_\text{trg}) - \mathbb E_{\mathbf s_i \in \mathcal D_S} (s_i) \Big),
\end{equation}
where $\mathbb E$ means expectation. Using the weighting strategy with $\alpha$ defined in \cref{eq:weight}, \clipdir{} is reformulated as
\begin{equation}
    \texttt{cs}\left( E(I_\text{edit}) - E(I_\text{src}), E(T_\text{trg}) -\mathbb E_{\mathbf s_i \in \mathcal D_S} \left( \alpha(\mathbf s_i) \cdot \mathbf s_i \right) \right).
\end{equation}
The same formulation applies for the target text $T_\text{trg}$ as well.

Across all configurations, with and without weighting, \augclip{} outperforms \clipdir{} in terms of alignment with human judgment and ground truth selection test accuracy. This emphasizes that our metric, \augclip{}, notably well-performs compared to \clipdir{}.

\subsection{Comparison with GPT-4V} \label{subsec:exp_gpt}

\begin{table}[ht]
    \centering
    \caption{\textbf{Comparison with GPT-4V.} We use \score{} for CelebA, EditVal, and Dreambooth, and \acc{} for TEdBench and MagicBrush.}
    \label{tab:exp_gpt}
    \resizebox{0.7\linewidth}{!}{%
    \begin{tabular}{lccccc}
    \toprule
     & CelebA & EditVal & DreamBooth & TEdBench & MagicBrush \\
    \midrule
    GPT-4V & 0.876 & \textbf{0.933} & 0.821 & \textbf{0.620} & 0.703 \\
    \augclip{} & \textbf{0.883} & 0.831 & \textbf{0.857} & 0.570 & \textbf{0.889} \\
    \bottomrule
    \end{tabular}
    }
\end{table}

Recently, GPT-4V \citep{gpt4v} has been employed in evaluating various tasks, including text-guided image editing, text-to-image generation, and image quality assessment. Since GPT-4V is one of the best-performing multi-modal large language models, we test the ability of GPT-4V’s effectiveness in evaluating the quality of text-guided edited images. For evaluation, we use the following prompt: ``Given a source image and two edited images, you should choose a better edited one based on the source and target text. Source text describes the source image, and target text describes the editing. A well-edited image should preserve the essence of the source image while following the target text.''

As shown in \cref{tab:exp_gpt}, GPT-4V outperforms \augclip{} in tasks such as EditVal and TEdBench, which involve simple edits like modifying a single object’s attribute. In contrast, our proposed metric, \augclip{}, effectively captures minor differences by augmenting attributes of the source image and target text and shows better performance in other benchmarks with complex scenarios. This finding is consistent with prior research \citep{zhang2023gpt}, which suggests that GPT-4V struggles to differentiate between images with subtle differences. 

\subsection{Additional Examples on Problem 1 of Directional CLIP Similarity}
\begin{figure}[ht]
    \centering
    \includegraphics[width=\linewidth]{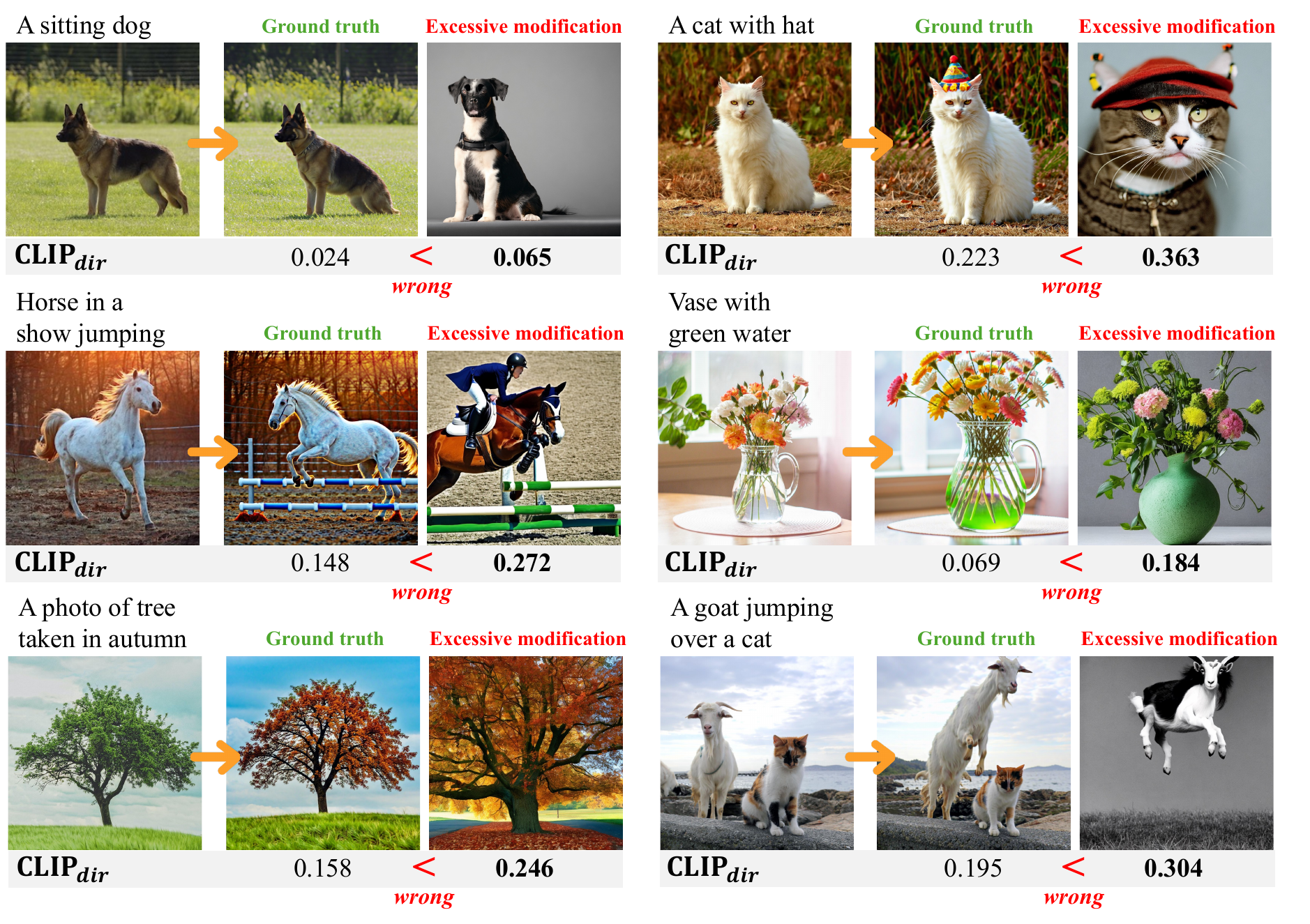}
    \caption{\textbf{Additional Examples on Problem 1 of Directional CLIP Similarity.} CLIP$_\text{dir}$ assigns higher scores to excessive modification, over well-edited ground truth images.}
    \label{fig:prob1_add}
\end{figure}
\clearpage

\subsection{Additional Examples on Problem 2 of Directional CLIP Similarity}
\begin{figure}[ht]
    \centering
    \includegraphics[width=\linewidth]{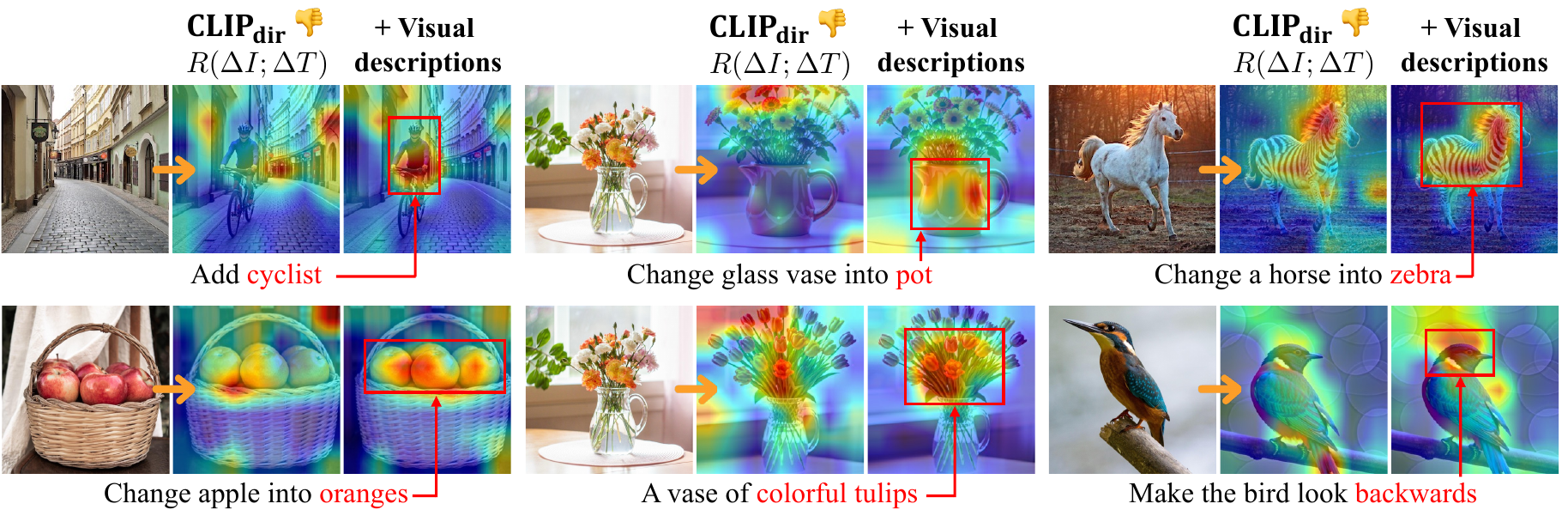}
    \caption{\textbf{Additional Examples on Problem 2 of Directional CLIP Similarity.} CLIP$_\text{dir}$ evaluates edited images by attending to irrelevant regions of the image. Adding visual annotations helps \clipdir{}  properly attend to edited regions.}
    \label{fig:prob2_add}
\end{figure}

To assess directional CLIP similarity's capability to focus on the image regions modified following the target text rather than peripheral or unchanged regions, we use the relevancy map \citep{chefer2021generic}, $\mR$. The relevancy map visualizes the transformer's attention on an image corresponding to a given text depending on their cosine similarity. Specifically, for an image $I\in \mathbb R^{h \times w}$ and text $T$, the relevancy map is computed as 

$$\mR(I;T) = \nabla_\mA \texttt{cs}(E(I),E(T);\mA) \odot \mA \in \mathbb{R}^{h \times w},$$
where $\mA$ represents the attention scores of the CLIP visual encoder and $\odot$ denotes the Hadamard product. To visualize the relevancy map of \clipdir{}, which is a cosine similarity between $\Delta I$ and $\Delta T$, we subtract the two relevancy maps as 

$$\mR(\Delta I; \Delta T) = \mR(I_\text{edit}; \Delta T) - \mR(I_\text{src}; \Delta T).$$

\cref{fig:clipdir_prob}(b) and \cref{fig:prob2_add} illustrate the relevancy maps of \clipdir{} across multiple cases and their improvement achieved by incorporating manually annotated visual descriptions. Unlike \clipdir{}, \augclip{} measures the cosine similarity between the estimated well-edited \textit{image} and the edited \textit{image}, rather than between an \textit{image} and \textit{text}. As a result, the relevancy map, which requires direct comparison of the image and text, cannot be applied to \augclip{}. 

\section{Qualitative Results} \label{sec:app_qual}

\begin{figure}[ht]
    \centering
    \includegraphics[width=\textwidth]{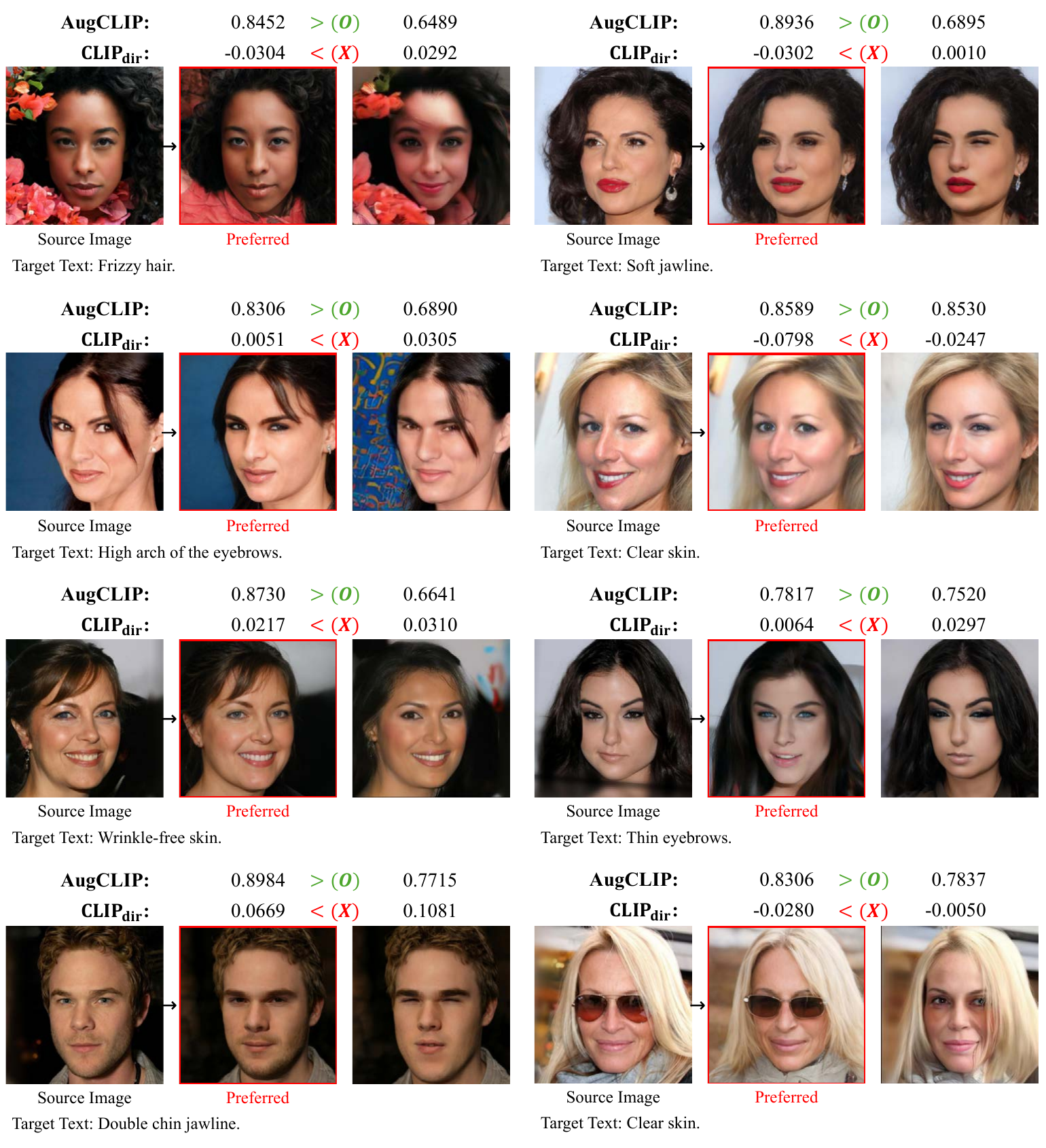}
    \caption{\textbf{Qualitative Results on CelebA (2AFC Test).}}
    \label{fig:qual_celeba}
\end{figure}

We present qualitative samples of the \textbf{2AFC Test}, as reported in \cref{tab:exp_main_2afc}, using the CelebA, EditVal, and DreamBooth datasets. For each dataset, we randomly select triplets consisting of a source image, target text, and edited images to demonstrate how \augclip{} consistently assigns higher scores to the edited image preferred by human evaluators. The preferred image, highlighted with a red box, appears in the middle. Each case represents a two-alternative forced choice (2AFC) survey, where the source image on the far left is altered into the middle and rightmost images. We observe that directional CLIP similarity often favors excessively modified images. For instance, in the second row of \cref{fig:qual_celeba}, where the target text is ``high arch of the eyebrows,'' directional CLIP similarity prefers an edited image that changes the gender of the source image into a man. Similarly, when the target text is ``wrinkle-free skin,'' directional CLIP similarity assigns a higher score to an image where the hair bangs are missing.

\newpage
\subsection{CelebA}
\clearpage

\subsection{EditVal}
\begin{figure}[ht]
    \centering
    \includegraphics[width=\textwidth]{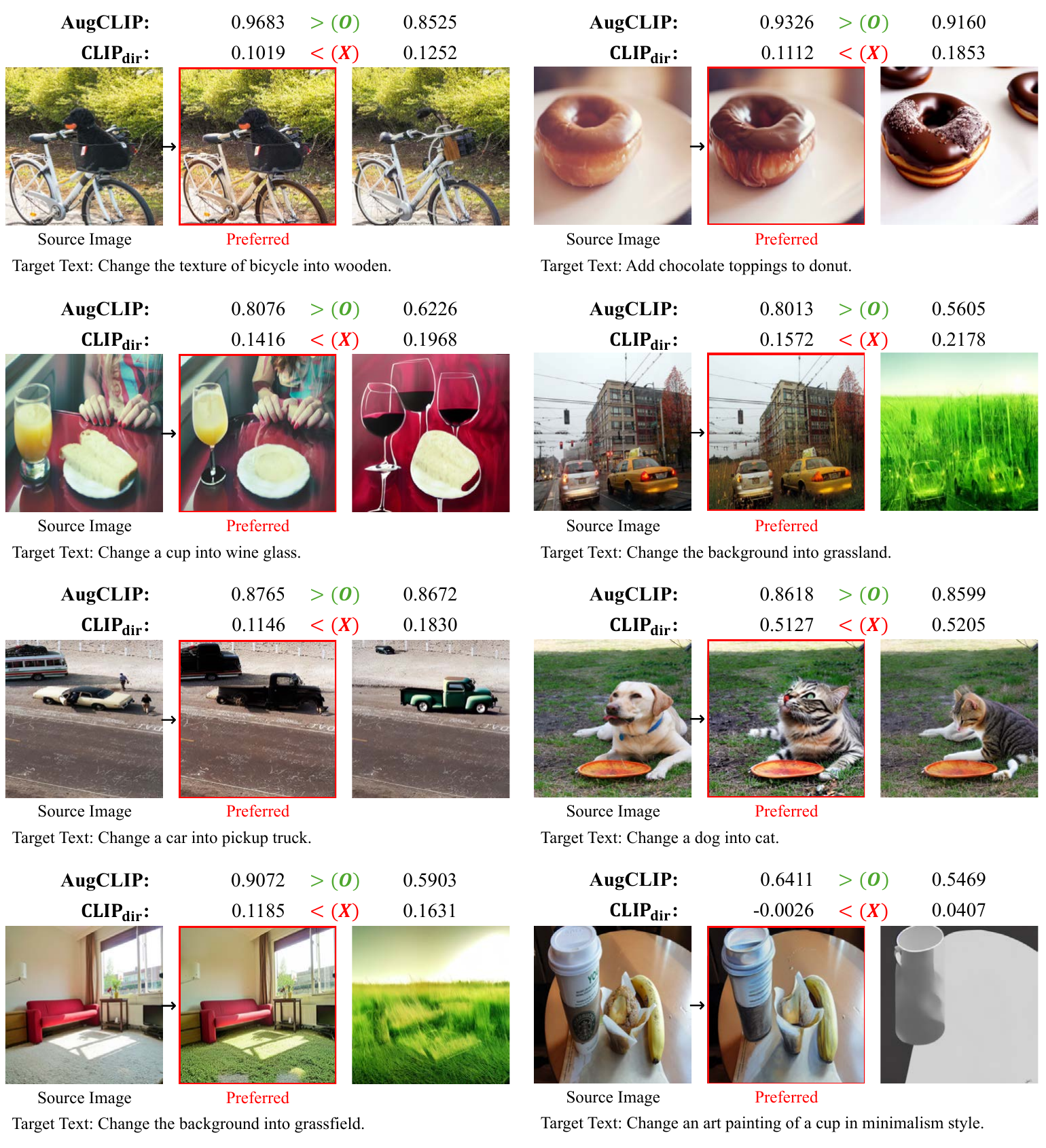}
    \caption{\textbf{Qualitative Results on EditVal (2AFC Test).}}
    \label{fig:qual_editval}
\end{figure}
\clearpage

\subsection{DreamBooth}
\begin{figure}[ht]
    \centering
    \includegraphics[width=\textwidth]{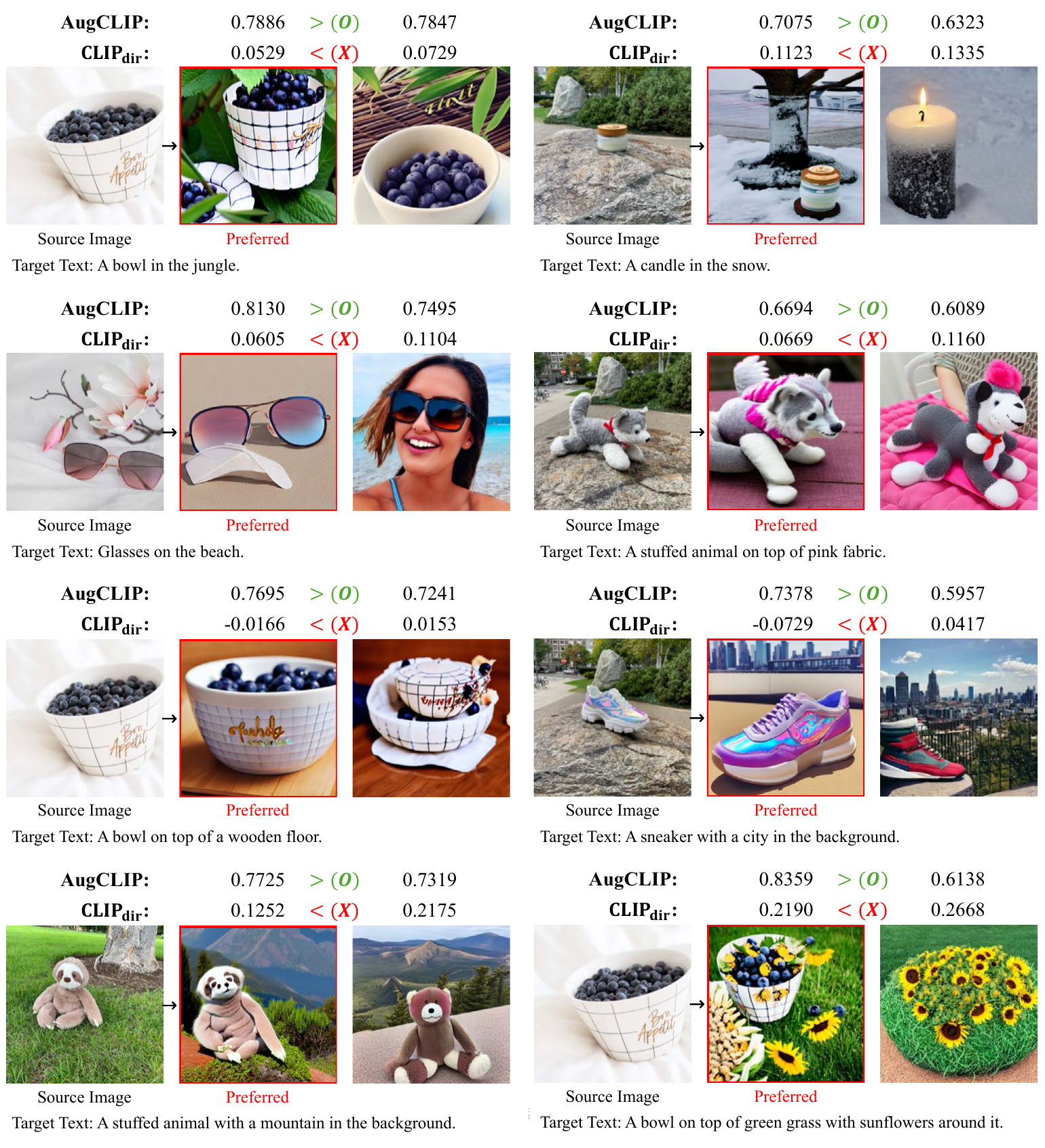}
    \caption{\textbf{Qualitative Results on DreamBooth dataset (2AFC test).}}
    \label{fig:qual_dreambooth}
\end{figure}
\clearpage

\begin{figure}[ht]
    \centering
    \includegraphics[width=0.8\textwidth]{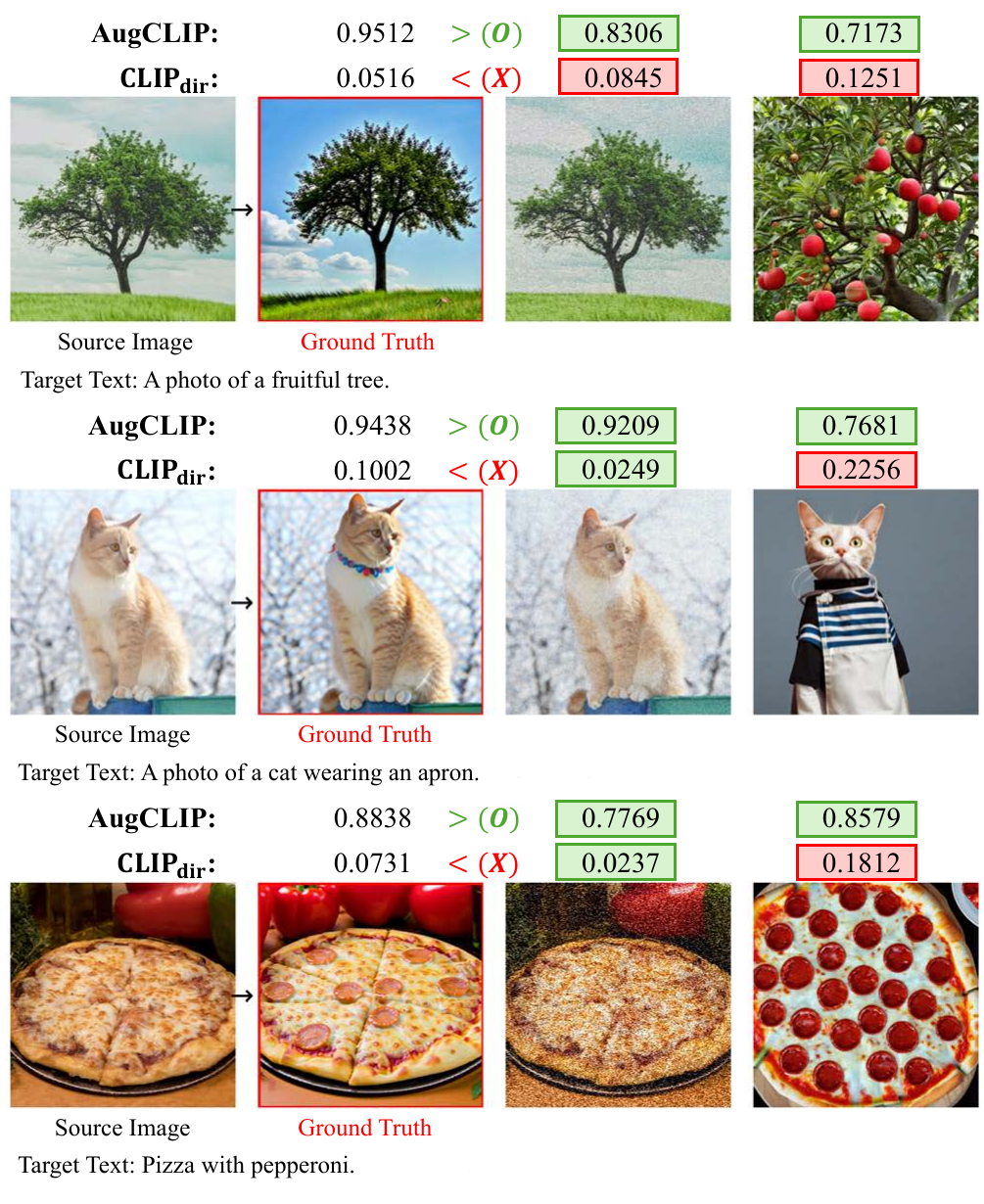}
    \caption{\textbf{Qualitative Results on TEdBench (Ground Truth Selection Test).}}
    \label{fig:qual_tedbench1}
\end{figure}

\begin{figure}[ht]
    \centering
    \includegraphics[width=0.8\textwidth]{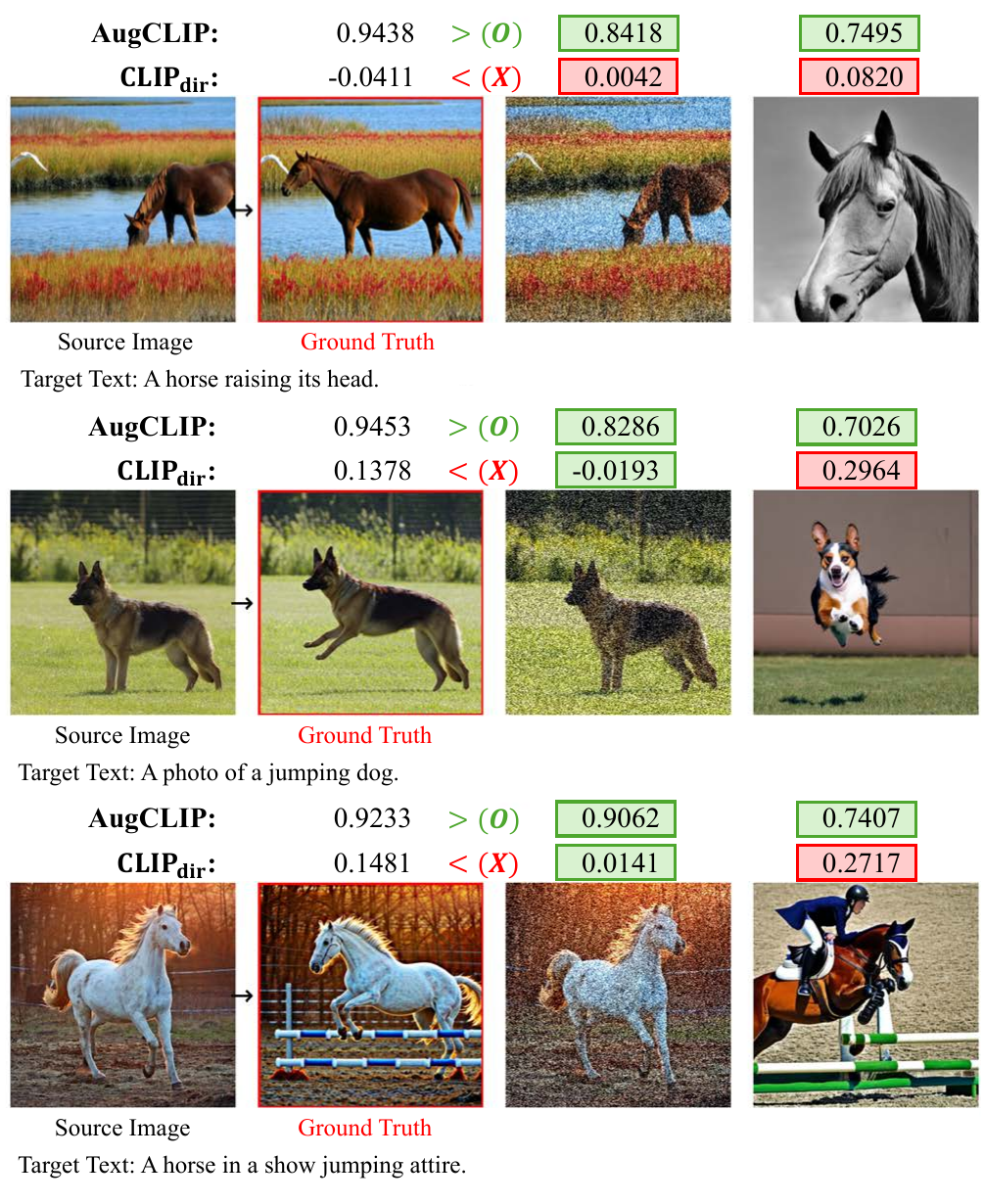}
    \caption{\textbf{Qualitative Results on TEdBench (Ground Truth Selection Test).}}
    \label{fig:qual_tedbench2}
\end{figure}

\begin{figure}[ht]
    \centering
    \includegraphics[width=0.8\textwidth]{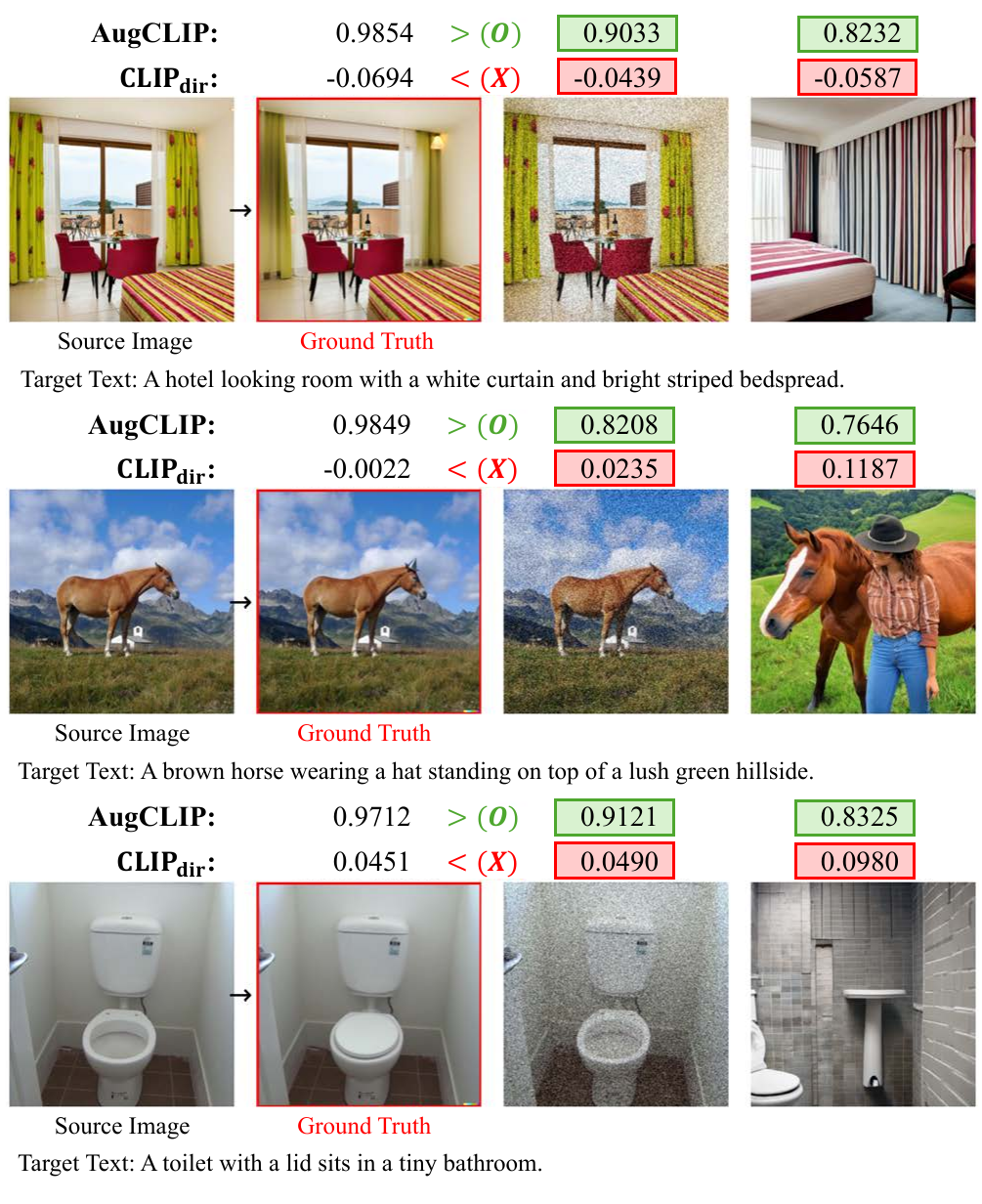}
    \caption{\textbf{Qualitative Results on MagicBrush (Ground Truth Selection Test).}}
    \label{fig:qual_magicbrush1}
\end{figure}

\begin{figure}[ht]
    \centering
    \includegraphics[width=0.8\textwidth]{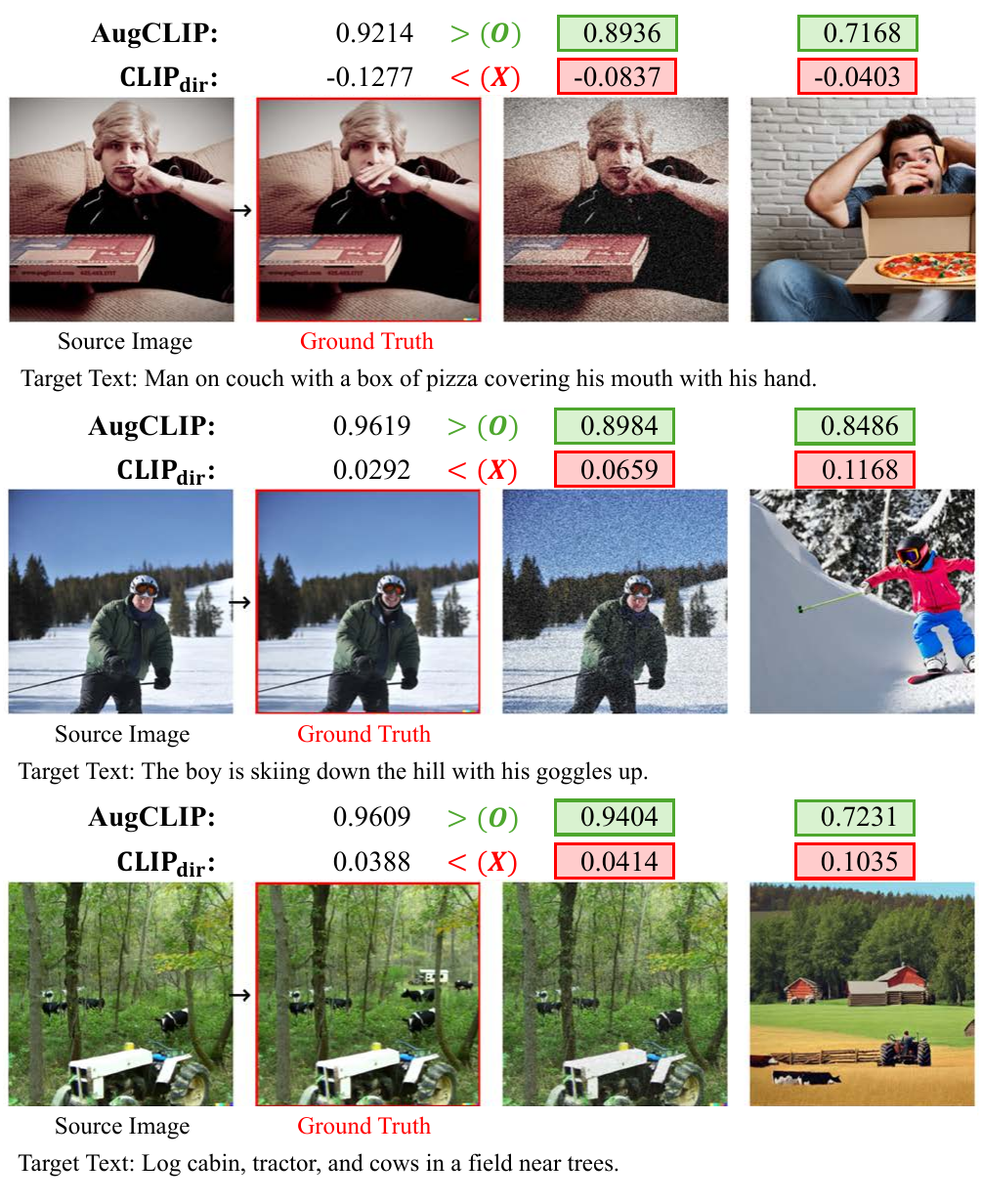}
    \caption{\textbf{Qualitative Results on MagicBrush (Ground Truth Selection Test).}}
    \label{fig:qual_magicbrush2}
\end{figure}

Additionally, we provide qualitative samples from the \textbf{Ground Truth Selection Test}, reported in \cref{tab:exp_main_gt}, using the TEdBench and MagicBrush datasets (\cref{fig:qual_tedbench1}, \ref{fig:qual_tedbench2} and \cref{fig:qual_magicbrush1}, \ref{fig:qual_magicbrush2}). In these cases, the ground truth image is located in the second column, the excessively preserved image in the third column, and the excessively modified image in the fourth column. Once again, we observe that directional CLIP similarity tends to prefer excessive modifications.

\subsection{TEdBench}
\clearpage

\clearpage

\subsection{MagicBrush}
\clearpage

\clearpage

\subsection{Failure Cases of \augclip{}}
\begin{figure}[ht]
    \centering
    \includegraphics[width=\linewidth]{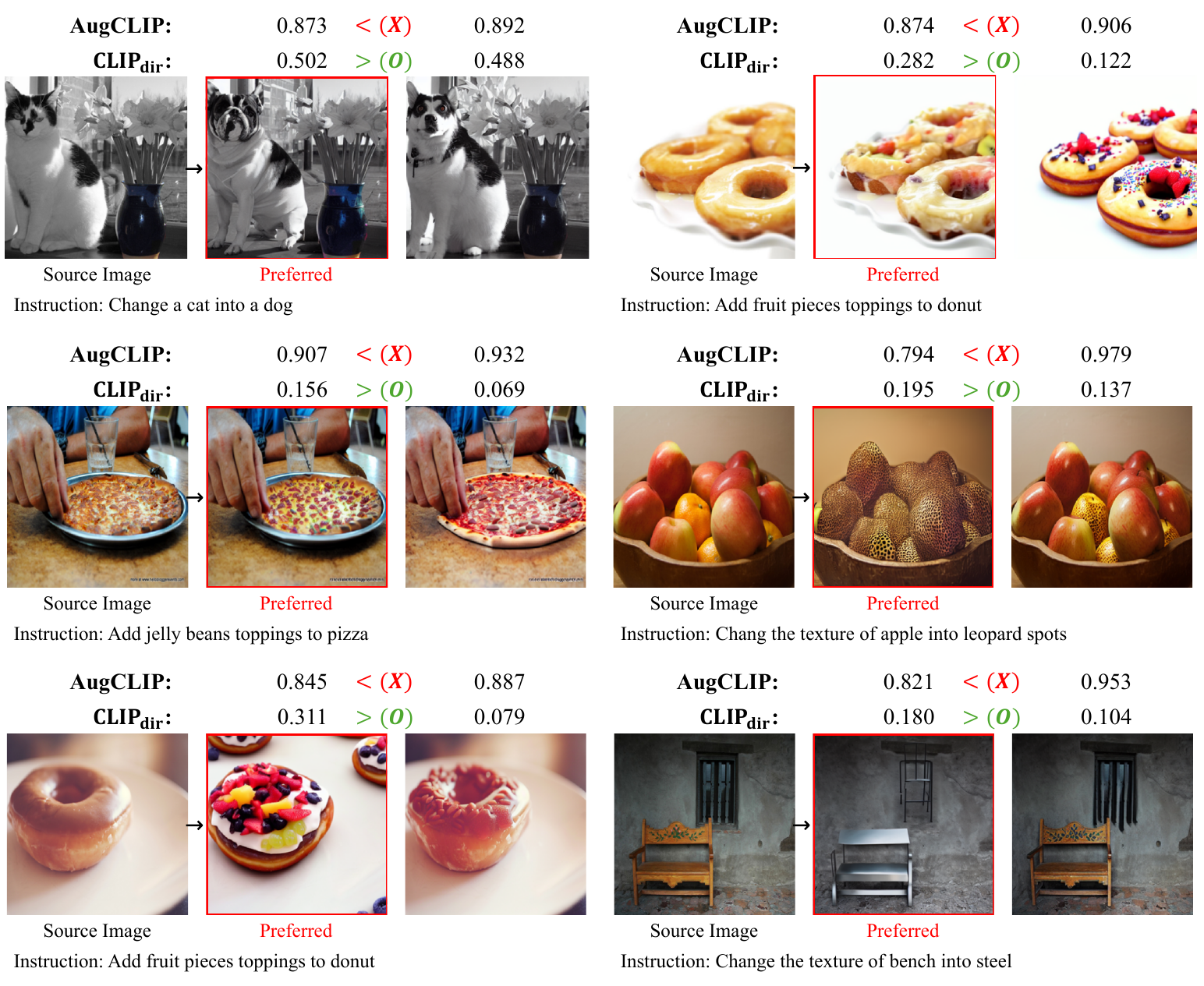}
    \caption{\black{\textbf{Failure Cases of \augclip{}.} Failure cases where directional CLIP similarity correctly assigns higher evaluation scores to images that human evaluators prefer, while \augclip{} fails to.}}
    \label{fig:limitation_failure}
\end{figure}

Compared to directional CLIP similarity, \augclip{} shows superior alignment with human evaluation and a stronger ability to classify ground truth images. However, there are several cases where directional CLIP similarity aligns closely with human preferences. \cref{fig:limitation_failure} illustrates examples where \augclip{} diverges from human judgment. 

For instance, in the first-row example, both edited images are adequately modified from the source to resemble the target text ``dog.'' However, the middle image emphasizes dog-like features more prominently while the right image exhibits subtler changes. Human evaluators tend to favor the more prominently modified one. In the example of adding fruit toppings to donuts, both edited images accurately depict fruit toppings while preserving the original content. Yet, human evaluators prefer the middle image, which better retains the original donut's color and texture. Here, preference is skewed toward better preservation.

Although the edits in these examples are well-executed in terms of balancing preservation and modification, human preferences remain inherently subjective and vary significantly from case to case. This highlights the limitation of evaluation metrics in fully capturing the nuances of human judgment.

\end{document}